\documentclass[sigconf]{acmart} 

\usepackage{balance}
\usepackage{mathrsfs}
\usepackage{enumitem} 
\usepackage{extarrows}
\usepackage{subcaption}
\usepackage{verbatim}
\usepackage{makecell}
\usepackage{diagbox}
\usepackage{bm}
\usepackage{multirow}
\usepackage{float}
\usepackage{placeins}
\usepackage[linesnumbered, ruled]{algorithm2e}
\SetAlFnt{\small}
\usepackage{epstopdf}
\usepackage{setspace}
\usepackage[super]{nth}
\usepackage{xcolor}
\usepackage{booktabs}
\usepackage{caption}
\usepackage{ragged2e}
\usepackage{tikz}
\usepackage[normalem]{ulem}
\usepackage{colortbl}
\usepackage[most]{tcolorbox}

\SetCommentSty{mycommfont}

\newbool{showcomments}
\booltrue{showcomments}
\newcommand{\cmt}[3]{%
    \ifbool{showcomments}{%
        {\color{#1}\textbf{#2}: #3}%
    }{}%
}

\allowdisplaybreaks[4]

\definecolor{mygray}{gray}{0.9}

\newcommand{\promptbox}[2]{%
\begin{center}
\fcolorbox{black}{mygray}{%
\begin{minipage}{0.96\linewidth}
\smallskip
\noindent\textbf{#1}\par
\smallskip
{\small\ttfamily #2}\par
\smallskip
\end{minipage}}
\end{center}
}

\tcbset{
  judgebox/.style={
    colback=gray!3,
    colframe=gray!50,
    boxrule=0.5pt,
    arc=2pt,
    left=6pt,right=6pt,top=6pt,bottom=6pt,
    breakable
  },
  spotcase/.style={
    enhanced,
    breakable,
    colback=white,
    colframe=black!20,
    boxrule=0.5pt,
    arc=2pt,
    left=6pt,right=6pt,top=4pt,bottom=4pt,
    boxsep=2pt,
    fonttitle=\bfseries,
    coltitle=black,
    attach title to upper,
    title={#1},
  }
}

\newcommand{\casebox}[2]{%
\begin{tcolorbox}[spotcase={#1}]
\ttfamily\footnotesize\raggedright
#2
\end{tcolorbox}
}

\begin{document}

\title{SPOT: Span-level Pause-of-Thought for Efficient and Interpretable Latent Reasoning in Large Language Models} 


\author{Yunlong	Chu}
\authornote{Both authors contributed equally to this research.}
\affiliation{%
  \institution{School of New Media and Communication, Tianjin University}
  \city{Tianjin}
  \country{China}}
\email{2024245030@tju.edu.cn}

\author{Minglai	Shao}
\authornotemark[1]
\affiliation{%
  \institution{School of New Media and Communication, Tianjin University}
  \city{Tianjin}
  \country{China}}
\email{shaoml@tju.edu.cn}

\author{Yuhang Liu}
\authornotemark[1]
\affiliation{%
  \institution{School of New Media and Communication, Tianjin University}
  \city{Tianjin}
  \country{China}
}
\email{liuyuhang_13@tju.edu.cn}

\author{Bing Hao}
\affiliation{%
  \institution{School of New Media and Communication, Tianjin University}
  \city{Tianjin}
  \country{China}}
\email{haobing@tju.edu.cn}

\author{Yumeng Lin}
\affiliation{%
  \institution{School of New Media and Communication, Tianjin University}
  \city{Tianjin}
  \country{China}}
\email{lym619@tju.edu.cn}

\author{Jialu Wang}
\affiliation{%
  \institution{Independent Contributor}
  \city{CA}
  \country{USA}}
\email{jwang470@ucsc.edu}

\author{Ruijie Wang}
\authornote{Corresponding author.} 
\affiliation{%
  \institution{School of Computer Science and Engineering, Beihang University}
  \city{Beijing}
  \country{China}}
\email{ruijiew@buaa.edu.cn}

\begin{abstract}
Explicit Chain-of-Thought (CoT) improves the reasoning performance of large language models (LLMs) but often incurs high inference cost due to verbose token-level traces.
While recent approaches reduce this overhead via concise prompting or step pruning, they largely \emph{truncate what the model says} rather than \emph{internalize what the model thinks}.
Latent reasoning offers a promising alternative by performing computation in the hidden space, yet prior methods face two critical challenges.
Many existing approaches rely on rigid point-to-point alignment, forcing a latent token to approximate the final representation of a reasoning step, which can be insufficient to capture the dense, variable-length semantics of an entire reasoning segment.
Furthermore, these methods often suffer from a lack of interpretability: latent states are commonly produced by unconstrained optimization or embedding mixing, yielding vectors that are difficult to decode or audit under the pretrained language head.
We propose \textbf{SPOT} (\textbf{S}pan-level \textbf{P}ause \textbf{O}f \textbf{T}hought), a flexible framework that compresses explicit CoT into compact latent \texttt{<pause>} tokens without enforcing a fixed response template.
At the core of SPOT is \textbf{Span-level Semantic Alignment}, a Sinkhorn optimal-transport objective that softly matches each \texttt{<pause>} token to the semantics of an entire reasoning segment, overcoming the rigidity of step-end alignment.
To further improve interpretability, SPOT introduces a \textbf{Frozen-Head Decoding Constraint} that keeps latent states directly decodable as token distributions under the frozen pretrained LM head, enabling readable keyword interpretations of latent thoughts.
Moreover, with two-stage training and no hard supervision on \texttt{<pause>}, SPOT enables controllable implicit reasoning intensity at inference.
Experiments on reasoning benchmarks demonstrate that SPOT improves accuracy by 2.3 points on average while reducing generated tokens by 37.5\% and provides faithful semantic interpretations of the latent reasoning process.
\end{abstract}



\maketitle

\renewcommand{\shortauthors}{xxx et al.}

\section{Introduction}\label{sec:intro}
Chain-of-Thought (CoT) prompting has become a fundamental paradigm for improving the reasoning performance of large language models (LLMs), by encouraging them to externalize intermediate steps in natural language~\cite{wei2022cot,kojima2022zeroshot,wang2023selfconsistency}.
Besides accuracy gains, explicit traces make the reasoning process more auditable---they allow practitioners to inspect intermediate claims, verify logical consistency, and diagnose failures~\cite{yao2022react,shinn2023reflexion}.
However, this transparency incurs prohibitive inference costs, often manifesting as “overthinking”---where models generate redundant steps that inflate computational overhead without proportional accuracy gains~\cite{sui2025stopoverthinking,aggarwal2025l1,chen2025efficientreasoning}.
Recent open reasoning models further highlight this tension between reasoning depth and computational efficiency~\cite{deepseek2025r1,qwen2024qwen25},  motivating methods that reduce reasoning overhead without giving up the benefits of structured reasoning.

To mitigate CoT overhead, a growing line of work studies \emph{efficient reasoning} by reducing the \emph{explicit} trajectory length.
Representative strategies include concise prompting, pruning low-utility steps, and dynamically allocating computation via early-exit or difficulty-aware control~\cite{aytes2025sot,wu2025litecot,deer2025dynamicexit,pear2025phaseentropy,wu2025cotpruning,wu2025cotpruning,tang2025concisehint,li2025stepentropy}. 
From the perspective of \emph{fast} versus \emph{slow} thinking, these methods primarily aim to reduce unnecessary \emph{slow} deliberation on easier instances, while still allowing multi-step reasoning behavior to emerge when the task demands it.
Nevertheless, these approaches remain subject to an intrinsic \emph{length--compute coupling} in autoregressive decoding, where inference-time computation is tied to the number of generated tokens.
Consequently, aggressively truncating the visible trace can also constrain the model's potential to perform multi-step deduction, which may lead to under-deliberation and lower accuracy on harder instances~\cite{li2025implicitreasoningsurvey}.

\begin{figure}[t]
  \centering
  \captionsetup{skip=3pt}
  \includegraphics[width=\linewidth]{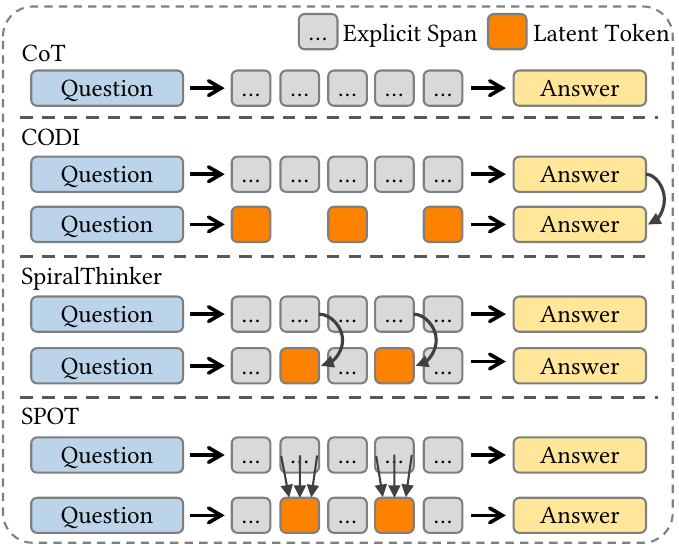} 
  \caption{Representative coupling paradigms in hybrid/latent reasoning.
  Each gray block denotes a contiguous explicit reasoning span (a paragraph-level segment delimited by blank lines), and each orange block denotes a latent token.}
  \label{fig:hybrid_paradigms}
\end{figure}

In contrast to fully constraining reasoning to the natural-language space, \textbf{implicit reasoning} internalizes intermediate inference into hidden states or latent tokens (special tokens whose hidden states carry intermediate computation)~\cite{li2025implicitreasoningsurvey,shen2025codi,cheng2024ccot,hao2024coconut}.
A key difficulty is \emph{drift}: when intermediate computation is carried primarily by latent trajectories, the model may deviate from the intended logical path while providing limited textual evidence for diagnosing intermediate decisions~\cite{piao2025spiralthinker,li2025implicitreasoningsurvey}.
To mitigate drift, recent implicit-reasoning methods further introduce \emph{explicit anchors} and interleave latent computation with selected text spans~\cite{piao2025spiralthinker,su2025tokenassorted,wang2025system15}.
Despite this progress, existing training paradigms for implicit reasoning still face several limitations.
First, many approaches adopt rigid point-to-point supervision, where each latent token is trained to match only an endpoint representation of a step or segment~\cite{shen2025codi,piao2025spiralthinker}.
Figure~\ref{fig:hybrid_paradigms} summarizes common distillation/alignment paradigms for learning implicit reasoning, where supervision is often imposed in a coarse one-to-one manner between latent tokens and explicit spans.
Such endpoint matching can under-represent the dense and variable-length semantics distributed across an entire reasoning span~\cite{zhang2025lightthinker,li2025implicitreasoningsurvey}.
Second, latent states are often hard to interpret: unconstrained latent optimization or embedding mixing may yield vectors that are poorly calibrated to the pretrained language head, making decoded “thoughts” ambiguous or unstable~\cite{deng2025latentsft,li2025implicitreasoningsurvey}.
Moreover, several frameworks impose a fixed interleaving template during training and inference, which reduces flexibility in how latent and explicit reasoning are composed across different instances and tasks~\cite{piao2025spiralthinker}.

To address these challenges, we propose \textbf{SPOT} (\textbf{S}pan-level \textbf{P}ause \textbf{O}f \textbf{T}hought), a framework that compresses explicit CoT into a small number of latent special tokens, denoted by \texttt{<pause>}, while retaining textual anchors and without enforcing a fixed interleaving template at inference time.
The key idea is to move beyond endpoint-based, point-to-point matching: as illustrated in Figure~\ref{fig:hybrid_paradigms}, SPOT performs \emph{span-level} alignment by coupling each \texttt{<pause>} token to the semantics of an entire variable-length reasoning span.
Concretely, SPOT introduces \textbf{Span-level Semantic Alignment} instantiated by a Sinkhorn-regularized optimal transport objective~\cite{cuturi2013sinkhorn,peyre2019cotbook}, which provides a structured coupling between a latent token and multiple teacher states within the associated span.
To improve interpretability, SPOT further enforces a \textbf{Frozen-Head Decoding Constraint} that keeps latent states compatible with the frozen pretrained LM head, enabling consistent token-level decoding of \texttt{<pause>} states without training auxiliary probes.
Finally, SPOT adopts a two-stage training procedure and supports inference-time control by externally inserting \texttt{<pause>} tokens, allowing practitioners to trade off accuracy and decoding length without committing to a fixed number or pattern of pauses.
We evaluate SPOT on four math reasoning benchmarks and GPQA-Diamond; on average across the five benchmarks, SPOT improves accuracy by 2.3 points while reducing the number of generated tokens by 37.5\% relative to the DeepSeek-R1-Distill-Qwen-7B backbone.

Our contributions can be summarized as follows:
\begin{itemize}[nosep, leftmargin=*]
    \item \textbf{Novel framework:} We propose \textbf{SPOT}, a flexible framework that compresses explicit CoT into compact latent \texttt{<pause>} tokens, enabling hybrid reasoning without enforcing a fixed response template.
    \item \textbf{Span-level alignment:} We introduce Span-level Semantic Alignment, instantiated with a Sinkhorn optimal-transport objective, to replace rigid point-to-point matching and robustly align latent tokens to variable-length reasoning spans.
    \item \textbf{Interpretable latent thoughts:} We propose a Frozen-Head Decoding Constraint that keeps latent states compatible with the frozen pretrained LM head, making latent thoughts directly decodable into readable keywords.
    \item \textbf{Extensive evaluation:} We conduct extensive experiments on five reasoning benchmarks, showing that SPOT improves accuracy while substantially reducing generation length, together with interpretability analyses that characterize the learned \texttt{<pause>} states.
\end{itemize}

\section{Background}\label{sec:2_background}

\subsection{Problem setting and notation.}
Given an input question $x$, a teacher model generates an explicit chain-of-thought (CoT) sequence $y=(y_1,\ldots,y_T)$, where $T$ is the token length.
We partition $y$ into $M$ consecutive reasoning spans $\{S_i\}_{i=1}^{M}$, where each span $S_i$ is a contiguous subsequence of tokens, and the set of spans forms a disjoint cover of $y$.
Let $|S_i|$ denote the length of span $S_i$, which may vary across spans and examples.

In this work, we focus on DeepSeek-R1-style CoT traces, in which reasoning steps are delimited by blank lines.
Accordingly, we adopt blank-line separators (\texttt{\textbackslash n\textbackslash n}) as \textbf{span boundaries}, consistent with prior research on efficient reasoning~\cite{li2025stepentropy,zhang2025lightthinker}.
The span granularity is a design choice and can be adjusted (e.g., paragraph- vs. sentence-level) depending on the target domain and output format.
In practice, to robustly detect these boundaries, we operate on the raw generated text (rather than token IDs) and treat \texttt{\textbackslash n\textbackslash n} as the canonical delimiter while also accepting common variants (e.g., extra spaces or multiple blank lines).
We apply span segmentation to the model’s reasoning segment; for DeepSeek-R1-style traces, this corresponds to the text enclosed by \texttt{<think>} and \texttt{</think>}.

Consider a decoder-only language model with a hidden dimension $d$, which produces a sequence of hidden states $H=(h_1,\ldots,h_T)$, where $h_t\in\mathbb{R}^{d}$ represents the hidden state at position $t$.
Let $V$ be the vocabulary size and $E\in\mathbb{R}^{V\times d}$ be the token embedding matrix. We denote the language modeling head (LM head) as $(W,b)$, where $W\in\mathbb{R}^{V\times d}$ is the weight matrix and $b\in\mathbb{R}^{V}$ is an optional bias vector.
Given a hidden state $h\in\mathbb{R}^{d}$, the model defines a token distribution as follows:
\begin{equation}
p(\cdot\mid h)=\mathrm{Softmax}(Wh+b).
\end{equation}
In subsequent sections, we utilize $E$ to map a vocabulary distribution to a vocabulary-induced soft embedding (Eq.~\eqref{eq:fh_soft}).

\subsection{Sinkhorn-Regularized Optimal Transport}
Optimal transport (OT) defines a principled measure of discrepancy between two discrete distributions over feature vectors.
Let $\{u_i\}_{i=1}^{m}$ and $\{v_j\}_{j=1}^{n}$ denote two sets of vectors, associated with probability weights $a\in\Delta^{m}$ and $b\in\Delta^{n}$, respectively.
Given a ground cost function $c(\cdot,\cdot)$, the cost matrix $C\in\mathbb{R}_{+}^{m\times n}$ is defined by $C_{ij}=c(u_i,v_j)$.

A transport plan $\Pi\in\mathbb{R}_{+}^{m\times n}$ quantifies the probability mass transported from $u_i$ to $v_j$.
The feasible set of couplings between $a$ and $b$ is given by:
\begin{equation}
\label{eq:ot_feasible}
\mathcal{U}(a,b)
=
\Bigl\{\Pi\in\mathbb{R}_{+}^{m\times n}:\ \Pi\mathbf{1}=a,\ \Pi^{\top}\mathbf{1}=b\Bigr\}.
\end{equation}
The unregularized OT distance is then defined as:
\begin{equation}
\label{eq:ot_value}
\mathrm{OT}(a,b;C)
=
\min_{\Pi\in\mathcal{U}(a,b)} \ \langle \Pi, C\rangle,
\qquad
\langle \Pi, C\rangle=\sum_{i,j}\Pi_{ij}C_{ij},
\end{equation}
which measures the minimum expected transport cost~\cite{peyre2019cotbook}.
To obtain a smoother objective that can be computed efficiently, entropic regularization leads to the Sinkhorn-regularized OT value~\cite{cuturi2013sinkhorn}:
\begin{equation}
\label{eq:sinkhorn_value}
\mathrm{OT}_{\varepsilon}(a,b;C)
=
\min_{\Pi\in\mathcal{U}(a,b)}
\ \langle \Pi, C\rangle
-\varepsilon \sum_{i,j}\Pi_{ij}\big(\log \Pi_{ij}-1\big),
\end{equation}
where $\varepsilon>0$ controls the regularization strength.
In practice, the value $\mathrm{OT}_{\varepsilon}(a,b;C)$ is computed via Sinkhorn iterations; implementation details are provided in Appendix~\ref{ap:ot_pseudocode}.

\begin{figure*}[t]
    \centering
    \captionsetup{skip=3pt}
    \includegraphics[width=1.0\textwidth]{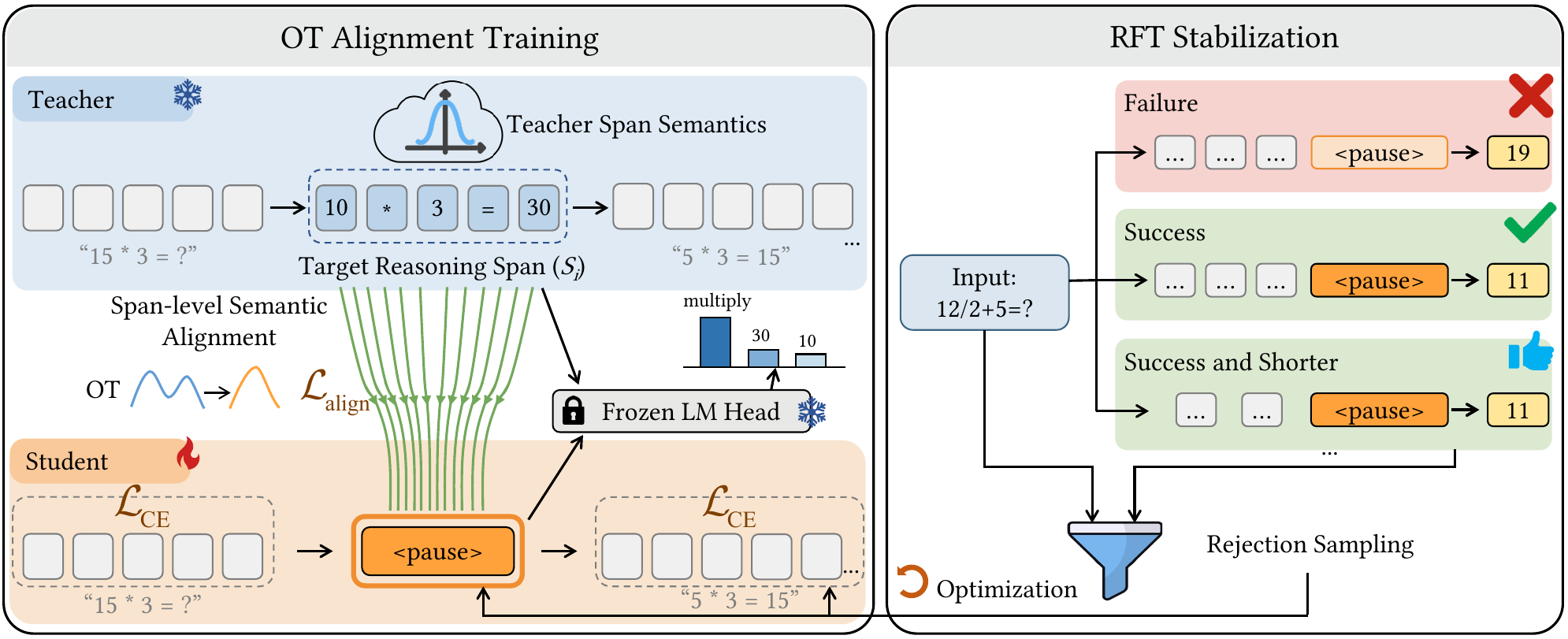}
    \caption{Overview of the SPOT framework. Stage I trains the student on SpanDrop sequences by anchoring \texttt{<pause>} hidden states to the corresponding teacher span under the frozen LM head, using a Sinkhorn-based alignment loss. Stage II applies RFT by selecting correct completions and preferring shorter ones, improving stability under external \texttt{<pause>} insertion.}
    \label{fig:spot_overview}
\end{figure*}

\section{The SPOT Framework}
\label{sec:3_framework}
In this section, we first provide an overview of the proposed \textbf{SPOT} framework, then introduce the technical details of the Frozen-Head Decoding Constraint and Span-level Semantic Alignment, and finally describe the two-stage training paradigm consisting of OT alignment training and rejection-sampled fine-tuning (RFT).

\subsection{Overview}
\label{sec:overview}
The SPOT framework (Figure~\ref{fig:spot_overview}) compresses chain-of-thought (CoT) traces into compact latent \texttt{<pause>} tokens. This approach effectively mitigates the computational overhead associated with verbose reasoning while keeping \texttt{<pause>} states decodable through the pretrained Language Modeling (LM) head.
We employ a teacher--student setup where the student shares the fixed teacher's backbone and undergoes a two-stage training process.
The procedure commences with \textbf{SpanDrop} data construction. This module partitions teacher CoT sequences using the span boundaries defined in Section~\ref{sec:2_background} and replaces selected spans with a single \texttt{<pause>}.
In Stage~I, we reuse the frozen pretrained LM head as a fixed projection for comparing student and teacher representations, and align each \texttt{<pause>} state to its paired teacher span using a Sinkhorn-based objective.
Stage~II performs \textbf{Rejection-Sampled Fine-Tuning (RFT)} by filtering sampled completions for correctness and preferring shorter ones, stabilizing implicit reasoning under diverse external \texttt{<pause>} insertion patterns.

\subsection{Training Data} \label{sec:data}
Given an input query $x$, the teacher generates an explicit CoT sequence
$y=(y_1,\ldots,y_T)$.
Consistent with the definition in Section~\ref{sec:2_background}, we partition $y$ into $M$ consecutive reasoning spans $\{S_i\}_{i=1}^{M}$, such that $y = S_1 \Vert \cdots \Vert S_M$.
To construct \textbf{SpanDrop} instances, we sample a subset of span indices $\mathcal{D}\subseteq\{1,\ldots,M\}$ by independently selecting each span with a probability of $0.3$.
For each index $i\in\mathcal{D}$, the corresponding span $S_i$ is replaced by a single special token \texttt{<pause>}, resulting in the compressed sequence $\tilde{y}$.
This substitution preserves the relative order of the remaining segments.
The \texttt{<pause>} token is integrated into the tokenizer vocabulary and shares the embedding matrix $E$ with standard tokens.
Consequently, each training instance constitutes a tuple $(x, y, \tilde{y})$, comprising the input, the complete teacher trace, and the span-compressed sequence. 

During training, the teacher model processes the full sequence $y$ to yield the final-layer hidden states
$H^{\text{tea}}=(h^{\text{tea}}_1,\ldots,h^{\text{tea}}_T)$.
For a specific span $S_i$, we aggregate the corresponding teacher states into: 
\begin{equation}
  H^{\text{tea}}(S_i)\;=\;\big[\,h^{\text{tea}}_t\,\big]_{y_t\in S_i}\in\mathbb{R}^{|S_i|\times d},
\end{equation}
which provides a variable-length representation of the span.

Simultaneously, the student model processes the span-dropped sequence $\tilde{y}$, conditioned on $x$.
For each replaced span $S_i$ ($i\in\mathcal{D}$), let $z_i\in\mathbb{R}^{d}$ denote the student's final-layer hidden state at the corresponding \texttt{<pause>} position.
The training process aligns the single latent state $z_i$ with the sequence of teacher states $H^{\text{tea}}(S_i)$ derived from the omitted span, establishing the supervision target for the alignment objective detailed in Section~\ref{sec:head_alignment}.
To ensure computational efficiency, we impose a maximum limit on the number of teacher states per span, employing uniform subsampling when $|S_i|$ exceeds this threshold (see Section~\ref{sec:exp_setup}).

\subsection{Frozen-Head Decoding and Span-level Semantic Alignment} \label{sec:head_alignment}

\subsubsection{Frozen-Head Decoding Constraint.}
To keep \texttt{<pause>} representations interpretable, SPOT employs a frozen-head projection that maps \texttt{<pause>} states into a fixed vocabulary-induced space used for alignment.
Given a hidden state $h\in\mathbb{R}^{d}$, we obtain a token distribution via the frozen LM head $(W,b)$:
\begin{equation}
  \label{eq:fh_prob}
  p(\cdot\mid h)=\mathrm{Softmax}(Wh+b).
\end{equation}
Subsequently, $h$ is mapped to a soft embedding within the token embedding space by computing the expectation over this distribution:
\begin{equation}
  \label{eq:fh_soft}
  \phi(h)=p(\cdot\mid h)^{\top}E \in \mathbb{R}^{d}.
\end{equation}

During Stage~I, the LM head $(W,b)$ and the embedding matrix $E$ remain frozen; consequently, $\phi(\cdot)$ constitutes a fixed vocabulary-induced mapping utilized throughout the alignment training process.
Furthermore, the frozen-head distribution $p(\cdot\mid h)$ enables the association of a hidden state with a concise set of high-probability vocabulary items.
Let $\mathrm{TopK}(p,K)$ denote the set of indices corresponding to the $K$ most probable vocabulary items under distribution $p$.
For a given hidden state $h$, we define 
\begin{equation}
\label{eq:topk_def}
\mathcal{T}_{K}(h)\;=\;\mathrm{TopK}\!\bigl(p(\cdot\mid h),K\bigr),
\end{equation}
and utilize $\mathcal{T}_{K}(h)$ as a compact lexical summary of the semantic content encoded by $h$ (with $K{=}20$ set as the default). 

\subsubsection{Span-level Semantic Alignment.}
For each dropped span $S_i$ ($i\in\mathcal{D}$), the student model generates a single hidden state $z_i\in\mathbb{R}^{d}$ at the corresponding \texttt{<pause>} position.
In contrast, the teacher model yields a sequence of token-level hidden states corresponding to the same span, denoted by $\{h^{\text{tea}}_t\}_{y_t\in S_i}$.
The objective is to align a single latent state with the semantics distributed across a variable-length span. 

This alignment is performed following the projection of both student and teacher states via the frozen-head mapping $\phi(\cdot)$ (Eq.~\eqref{eq:fh_soft}), thereby embedding them into a shared semantic space.
Specifically, the student state is mapped to 
\begin{equation}
\tilde{z}_i = \phi(z_i) \in \mathbb{R}^{d},
\end{equation}
while each teacher token state is transformed into $\tilde{h}^{\text{tea}}_t=\phi(h^{\text{tea}}_t)$.
Aggregating the projected teacher states yields the matrix: 
\begin{equation}
\tilde{H}^{\text{tea}}(S_i)
=
\big[\,\tilde{h}^{\text{tea}}_t\,\big]_{y_t\in S_i}
\in \mathbb{R}^{|S_i|\times d}.
\end{equation}
To softly match a single \texttt{<pause>} state to a variable-length span, we use the Sinkhorn-regularized OT objective in Eq.~\eqref{eq:sinkhorn_value} as a differentiable token-level set-matching loss between $\tilde{z}_i$ and $\{\tilde{h}^{\text{tea}}_t\}_{y_t\in S_i}$.
Given that the source distribution possesses a single support point, this one-to-span OT formulation simplifies to a \emph{weighted point-to-set distance}—effectively distributing the unit mass from $\tilde{z}_i$ across the span tokens according to the target marginal—rather than a general many-to-many transport plan.
Concretely, the ground cost between $\tilde{z}_i$ and each projected teacher token state is defined as follows: 
\begin{equation}
\label{eq:ot_cost}
C_i \in \mathbb{R}_{+}^{1\times |S_i|},
\qquad
(C_i)_{1,t}
=
\bigl\lVert
\tilde{z}_i-\tilde{h}^{\text{tea}}_t
\bigr\rVert_2^2,
\quad y_t\in S_i .
\end{equation}
The source marginal concentrates all probability mass on the single \texttt{<pause>} state, whereas the target marginal assigns uniform mass across the tokens within the span:
\begin{equation}
a_i = [1] \in \Delta^{1},
\qquad
b_i = \frac{1}{|S_i|}\mathbf{1} \in \Delta^{|S_i|}.
\end{equation}
Finally, the span-level alignment loss is computed as the average of the Sinkhorn-regularized OT values over all dropped spans:
\begin{equation}
\label{eq:spot_align}
\mathcal{L}_{\text{align}}
=
\frac{1}{|\mathcal{D}|}
\sum_{i\in\mathcal{D}}
\mathrm{OT}_{\varepsilon}\!\left(a_i,\, b_i;\, C_i\right),
\end{equation}
where $\mathrm{OT}_{\varepsilon}(a,b;C)$ denotes the optimal value of Eq.~\eqref{eq:sinkhorn_value}.
In this formulation, the alignment leverages \emph{all} token states within the teacher span, thereby providing richer supervision compared to endpoint-only matching or simple pooling-based regression, while maintaining compatibility with variable span lengths.

\subsection{Training Paradigm}
\label{sec:training}
SPOT is trained in two stages.
Stage~I focuses on learning span-aligned latent representations for the \texttt{<pause>} token using SpanDrop data and an Optimal Transport (OT)-based alignment objective.
Stage~II employs Rejection-Sampled Fine-Tuning (RFT) to enhance the robustness of the model to externally inserted \texttt{<pause>} tokens during inference. 

\subsubsection{Stage I: OT alignment training.}
Given a SpanDrop sequence $\tilde{y}=(\tilde{y}_1,\ldots,\tilde{y}_{\tilde{T}})$, we optimize the student model in Stage~I using a composite objective comprising (i) an autoregressive next-token prediction loss on observed tokens and (ii) the span-level alignment loss defined in Eq.~\eqref{eq:spot_align}.
In Stage~I, we freeze the pretrained LM head $(W,b)$ and the token embedding matrix $E$, and optimize the remaining student parameters $\theta$ with the objective below.
The next-token prediction loss is applied exclusively to non-\texttt{<pause>} tokens: 
\begin{equation}
  \label{eq:stage1_CE}
  \mathcal{L}_{\text{CE}}
  =
  -\sum_{t=1}^{\tilde{T}}
  \mathbf{1}[\tilde{y}_t\neq \texttt{<pause>}]\,
  \log p_{\theta}\!\left(\tilde{y}_t \mid x,\tilde{y}_{<t}\right).
\end{equation}
The indicator function $\mathbf{1}[\tilde{y}_t\neq \texttt{<pause>}]$ masks the \texttt{<pause>} positions, thereby preventing the cross-entropy term from imposing token-level supervision on the \texttt{<pause>} token itself.
We combine $\mathcal{L}_{\text{CE}}$ with the alignment loss $\mathcal{L}_{\text{align}}$ (Eq.~\eqref{eq:spot_align}) as follows:
\begin{equation}
  \label{eq:stage1_total}
  \mathcal{L}_{\text{stage1}}
  =
  \mathcal{L}_{\text{CE}}
  +\lambda\,\mathcal{L}_{\text{align}},
\end{equation}
where $\lambda$ controls the relative strength of span-level alignment.

Crucially, Stage~I imposes no cross-entropy supervision at \texttt{<pause>} positions; rather, the \texttt{<pause>} token is introduced via the SpanDrop construction and serves as part of the conditioning context. Its hidden state is shaped solely through the span-level alignment objective.
Consequently, the model is not trained to emit \texttt{<pause>} according to a fixed template. Instead, it learns to utilize \texttt{<pause>} as a carrier of internal computation, allowing inference-time insertion patterns to be specified externally to adjust the degree of implicit reasoning.

\subsubsection{Stage II: RFT stabilization.}
Although Stage~I induces span-aligned \texttt{<pause>} representations, the trained model may remain sensitive to the specific placement and density of \texttt{<pause>} tokens inserted externally during inference.
Stage~II implements Rejection-Sampled Fine-Tuning (RFT) to improve robustness against such variations.

For each input query $x$, we first generate an uncompressed reference completion $y^{(0)}$ that contains no \texttt{<pause>} tokens. Subsequently, we sample a set of candidate completions $\{y^{(k)}\}_{k=1}^{K_{\text{cand}}}$ by inserting \texttt{<pause>} tokens according to various predefined patterns.
A rejection step filters out candidates that yield incorrect final answers.
Among the remaining correct candidates, we prioritize shorter outputs based on the scoring function:\begin{equation}
  \label{eq:rft_score}
  s\!\left(y^{(k)};x\right)
  =
  1-\frac{|y^{(k)}|}{|y^{(0)}|},
\end{equation}
where $|\cdot|$ denotes the output length. The candidate with the highest score is selected as the RFT target $\hat{y}$.

The model is then fine-tuned on $\hat{y}$ using the same next-token prediction loss:
\begin{equation}
  \label{eq:rft_ce}
  \mathcal{L}_{\text{RFT}}
  =
  -\sum_{t=1}^{|\hat{y}|}
  \mathbf{1}[\hat{y}_t\neq \texttt{<pause>}]\,
  \log p_{\theta}\!\left(\hat{y}_t \mid x,\hat{y}_{<t}\right).
\end{equation}
Consistent with Stage~I, the indicator $\mathbf{1}[\hat{y}_t\neq\texttt{<pause>}]$ ensures that the cross-entropy loss does not treat \texttt{<pause>} as a prediction target.
However, Stage~II explicitly allows the optimization of the embedding vector for \texttt{<pause>}, which empirically improves the stability of generations conditioned on externally inserted \texttt{<pause>} tokens.
During RFT, we keep $(W,b)$ frozen and freeze $E$ except for the \texttt{<pause>} embedding. All remaining trainable parameters are the same $\theta$ optimized in Stage~I.

\subsection{Inference-time insertion.} \label{sec:inference}
During inference, we externally inject \texttt{<pause>} tokens into the reasoning segment (for DeepSeek-style outputs, the \texttt{<think>}...\texttt{</think>} region) at intervals of $N$ spans (defined in Section~\ref{sec:2_background}), and never insert \texttt{<pause>} outside this segment.
This mechanism affords explicit control over the extent to which intermediate computation is performed implicitly, while preserving the native response format of the model.
SPOT also supports more flexible insertion rules; see Appendix~\ref{ap:insertion} for further discussion.

\section{Experiments} \label{sec:5_experiment}
This section evaluates \textbf{SPOT} on math reasoning benchmarks and an out-of-domain (OOD) science QA benchmark.
We first describe the experimental setup, including training data, evaluation protocol, baselines, and implementation details.
We then report the main results, followed by ablation studies and interpretability analyses.

\subsection{Experimental Setup}\label{sec:exp_setup}

\subsubsection{Backbone and Training Data}
All methods adopt DeepSeek-R1-Distill-Qwen-7B~\cite{deepseek2025r1} as the backbone for all methods.
SPOT is trained on the GSM8K training set~\cite{cobbe2021gsm8k}.

\subsubsection{Benchmarks}
Benchmarks include GSM8K~\cite{cobbe2021gsm8k} and three harder math evaluations: MATH500~\cite{hendrycks2021math}, AIME 2024, and AIME 2025~\cite{aime}.
Out-of-domain generalization is assessed on GPQA-Diamond~\cite{rein2023gpqa}, a science QA benchmark that differs substantially from GSM8K-style grade-school math.
We summarize the dataset sizes and answer-format constraints in Table~\ref{tab:benchmark_stats} (Appendix~\ref{ap:benchmark_details}).

\subsubsection{Evaluation Protocol}
Prompt templates and task-specific answer format constraints are provided in Appendix~\ref{sec:appendix_prompts}.
Decoding uses temperature $0.6$, top-$p$ $0.95$, and a maximum generation length of 16{,}384 tokens.
All reported results are averaged over ten decoding runs with different random seeds.
In the main results, we use $N{=}3$ on all math benchmarks and $N{=}1$ on GPQA-Diamond,
where $N$ is the number of explicit spans between two consecutive inference-time \texttt{<pause>} insertions (Section~\ref{sec:inference}).

\subsubsection{Evaluation Metrics}
\textbf{Pass@1 accuracy} ($\mathbf{Acc}$) is computed from \emph{one} sampled completion per instance; we repeat decoding with 10 random seeds and report the mean.
\textbf{Output length} is reported as $\mathbf{\#L}$, defined as the total number of generated tokens in the full response (reasoning plus final answer) under the model tokenizer.

\subsubsection{Baselines} \label{sec:baselines}
We compare SPOT with representative baselines under the same DeepSeek-R1-Distill-Qwen-7B backbone.
\textbf{Vanilla} denotes the original model without intervention.
We include explicit trace-control baselines (CCoT~\cite{ccot}, ConciseHint~\cite{tang2025concisehint}, Step Entropy~\cite{li2025stepentropy}, L1-Max~\cite{aggarwal2025l1}, DEER~\cite{deer2025dynamicexit}) and implicit/latent baselines (COCONUT~\cite{hao2024coconut}, CODI~\cite{shen2025codi}, LightThinker~\cite{zhang2025lightthinker}, Latent-SFT~\cite{deng2025latentsft}).
Detailed descriptions are in Appendix~\ref{sec:appendix_baselines}.

\subsection{Implementation Details}\label{sec:impl_details}
Experiments are run on a single machine with $2\times$ NVIDIA RTX 5880 Ada GPUs (48\,GB each) and dual-socket Intel Xeon Gold 6454S CPUs.
Training uses PyTorch with the HuggingFace Transformers framework and Distributed Data Parallel (DDP) across two GPUs.
We train the model with bfloat16 precision.

\noindent \textbf{Training configuration.}
Stage~I uses a maximum sequence length of 4096, per-device batch size 1, and gradient accumulation of 8.
Optimization uses AdamW for 5 epochs with learning rate $2\times 10^{-5}$, warmup ratio 0.05, and max gradient norm 1.0.
For Sinkhorn OT, we set \texttt{blur}=0.05 and \texttt{scaling}=0.9.
Unless stated otherwise, we set $\lambda=1.0$ (Eq.~\ref{eq:stage1_total}), and each teacher span is capped at 256 token states with uniform subsampling when exceeded.
Stage~II (RFT) follows the same optimizer settings.

\noindent \textbf{Parameter-efficient adaptation.}
In both stages, we fine-tune the student model using LoRA while keeping the teacher model frozen.
We use LoRA rank $r{=}64$, scaling $\alpha{=}128$, and dropout 0.1, applied to the linear projections in the attention blocks ($q,k,v,o$) and the MLP layers (gate/up/down).
The LM head is kept frozen throughout; in Stage~II, we additionally unfreeze the embedding vector of \texttt{<pause>} while keeping all other token embeddings frozen.

\newcommand{\cell}[2]{#1\raisebox{-0.35ex}{\tiny \textcolor{black!70}{#2}}}

\newcommand{\cellgray}[2]{#1\raisebox{-0.35ex}{\scriptsize \textcolor{black!70}{#2}}}

\begin{table*}[htbp]
\centering
\setlength{\tabcolsep}{2pt}
\renewcommand{\arraystretch}{1.2} 
\begin{tabular}{lcccccccccccc}
\toprule
\multirow{2}{*}{\textbf{Method}}
& \multicolumn{2}{c}{\textbf{GSM8K}}
& \multicolumn{2}{c}{\textbf{MATH500}}
& \multicolumn{2}{c}{\textbf{AIME 2024}}
& \multicolumn{2}{c}{\textbf{AIME 2025}}
& \multicolumn{2}{c}{\textbf{GPQA-Diamond}}
& \multicolumn{2}{c}{\textbf{Average}} \\ 
\cmidrule(lr){2-3}\cmidrule(lr){4-5}\cmidrule(lr){6-7}\cmidrule(lr){8-9}\cmidrule(lr){10-11}\cmidrule(lr){12-13}
& Acc(\%) & \#L
& Acc(\%) & \#L
& Acc(\%) & \#L
& Acc(\%) & \#L
& Acc(\%) & \#L
& \textbf{$\Delta$Acc} & \textbf{$\Delta$\#L} \\ 
\midrule
Vanilla
& 89.61 & 1,316
& 92.40 & 3,775
& \textbf{54.33} & 10,874
& 36.00 & 11,357
& 50.10 & 8,238
& -- & -- \\
\midrule

CCoT
& \cell{89.83}{+0.2\%} & \cell{1,226}{-6.9\%}
& \cell{92.26}{-0.1\%} & \cell{3,133}{-17.0\%}
& \cell{52.33}{-2.0\%} & \cell{9,884}{-9.1\%}
& \cell{35.33}{-0.7\%} & \cell{10,842}{-4.5\%}
& \cell{50.83}{+0.7\%} & \cell{7,869}{-4.5\%} 
& -0.4\% & -8.4\% \\

DEER
& \cell{91.05}{+1.4\%} & \cell{985}{-25.2\%}
& \cell{92.60}{+0.2\%} & \cell{2,486}{-34.1\%}
& \cell{52.67}{-1.7\%} & \cell{9,306}{-14.4\%}
& \cell{\underline{35.67}}{-0.3\%} & \cell{10,124}{-10.9\%}
& \cell{50.65}{+0.6\%} & \cell{5,323}{-35.4\%} 
& +0.0\% & -24.0\% \\

ConciseHint
& \cell{89.86}{+0.3\%} & \cell{872}{-33.7\%}
& \cell{92.62}{+0.2\%} & \cell{2,447}{-35.2\%}
& \cell{\underline{53.33}}{-1.0\%} & \cell{9,053}{-16.8\%}
& \cell{34.67}{-1.3\%} & \cell{9,822}{-13.5\%}
& \cell{\underline{52.99}}{+2.9\%} & \cell{7,090}{-13.9\%} 
& +0.2\% & -22.6\% \\

Step Entropy
& \cell{90.46}{+0.9\%} & \cell{1,269}{-3.6\%}
& \cell{\underline{92.80}}{+0.4\%} & \cell{2,649}{-29.8\%}
& \cell{\underline{53.33}}{-1.0\%} & \cell{8,567}{-21.2\%}
& \cell{34.67}{-1.3\%} & \cell{9,637}{-15.2\%}
& \cell{51.98}{+1.9\%} & \cell{7,167}{-13.0\%} 
& +0.2\% & -16.6\% \\

L1-Max
& \cell{90.48}{+0.9\%} & \cell{1,448}{+10.0\%}
& \cell{89.40}{-3.0\%} & \cell{2,580}{-31.7\%}
& \cell{30.33}{-24.0\%} & \cell{3,866}{-64.5\%}
& \cell{24.00}{-12.0\%} & \cell{4,636}{-59.2\%}
& \cell{47.15}{-3.0\%} & \cell{5,666}{-31.2\%} 
& -8.2\% & -35.3\% \\
\midrule

COCONUT
& \cell{70.36}{-19.3\%} & \cell{385}{-70.7\%}
& \cell{69.88}{-22.5\%} & \cell{494}{-86.9\%}
& \cell{8.67}{-45.7\%} & \cell{963}{-91.1\%}
& \cell{6.67}{-29.3\%} & \cell{1,135}{-90.0\%}
& \cell{20.35}{-29.8\%} & \cell{989}{-88.0\%} 
& -29.3\% & -85.4\% \\

CODI
& \cell{79.39}{-10.2\%} & \cell{458}{-65.2\%}
& \cell{76.60}{-15.8\%} & \cell{624}{-83.5\%}
& \cell{10.67}{-43.7\%} & \cell{1,237}{-88.6\%}
& \cell{6.33}{-29.7\%} & \cell{1,836}{-83.8\%}
& \cell{23.65}{-26.5\%} & \cell{1,265}{-84.6\%} 
& -25.2\% & -81.2\% \\

LightThinker
& \cell{90.55}{+0.9\%} & \cell{1,166}{-11.4\%}
& \cell{91.42}{-1.0\%} & \cell{3,064}{-18.9\%}
& \cell{38.67}{-15.7\%} & \cell{8,354}{-23.2\%}
& \cell{29.67}{-6.3\%} & \cell{9,543}{-16.0\%}
& \cell{50.54}{+0.4\%} & \cell{7,063}{-14.3\%} 
& -4.3\% & -16.7\% \\

Latent-SFT
& \cell{88.42}{-1.2\%} & \cell{457}{-65.3\%}
& \cell{79.80}{-12.6\%} & \cell{787}{-79.2\%}
& \cell{20.33}{-34.0\%} & \cell{1,537}{-85.9\%}
& \cell{10.67}{-25.3\%} & \cell{2,074}{-81.7\%}
& \cell{34.65}{-15.5\%} & \cell{1,831}{-77.8\%} 
& -17.7\% & -78.0\% \\

\midrule

\rowcolor{mygray}
SPOT-stage1
& \cellgray{\underline{92.19}}{+2.6\%} & \cellgray{576}{-56.3\%}
& \cellgray{92.24}{-0.2\%} & \cellgray{1,626}{-56.9\%}
& \cellgray{52.67}{-1.7\%} & \cellgray{8,281}{-23.9\%}
& \cellgray{34.67}{-1.3\%} & \cellgray{9,471}{-16.6\%}
& \cellgray{51.01}{+0.9\%} & \cellgray{4,142}{-49.7\%} 
& +0.1\% & -40.7\% \\

\rowcolor{mygray}
SPOT-stage2
& \cellgray{\textbf{92.72}}{+3.1\%} & \cellgray{630}{-52.1\%}
& \cellgray{\textbf{93.80}}{+1.4\%} & \cellgray{2,154}{-43.0\%}
& \cellgray{\underline{53.33}}{-1.0\%} & \cellgray{7,885}{-27.5\%}
& \cellgray{\textbf{39.33}}{+3.3\%} & \cellgray{9,568}{-15.8\%}
& \cellgray{\textbf{54.55}}{+4.5\%} & \cellgray{4,181}{-49.3\%} 
& \textbf{+2.3\%} & -37.5\% \\

\bottomrule

\end{tabular}
\caption{\textbf{Main Results on DeepSeek-R1-Distill-Qwen-7B.} Results are averaged over 10 random seeds. Each cell reports the absolute metric value, with subscripts showing the difference vs. \textbf{Vanilla} (Acc: absolute percentage points; Length: relative \%). Best and second-best results are in \textbf{bold} and \underline{underlined}, respectively.}
\label{tab:main_results}
\end{table*}

\subsection{Main Results}
As shown in Table~\ref{tab:main_results}, SPOT substantially shortens generation while matching or exceeding the accuracy of the Vanilla backbone.
By contrast, existing efficiency baselines often yield only modest length reductions, and their accuracy degradation becomes more apparent on harder benchmarks.
Specifically, SPOT reduces output length by \textbf{52.1\%} on GSM8K and \textbf{43.0\%} on MATH500 while improving accuracy by \textbf{+3.1} and \textbf{+1.4} points, respectively.
The improvements persist on more challenging and out-of-domain settings: On the harder AIME2025 benchmark, SPOT-stage2 achieves \textbf{39.33\%} accuracy (+3.3 points) while generating \textbf{15.8\%} fewer tokens, and it generalizes to out-of-domain GPQA-Diamond with \textbf{54.55\%} accuracy (+4.5 points) alongside a \textbf{49.3\%} reduction in output length.
Overall, SPOT achieves strong compression without the pronounced accuracy drop often observed in aggressive explicit shortening or fully latent compression baselines, and Stage~II further stabilizes this behavior, yielding the best results among SPOT variants.

\subsection{Ablation Studies}
This section analyzes SPOT from four complementary aspects: (i) controllability under inference-time external \texttt{<pause>} insertion, (ii) the design of the alignment objective, (iii) the compression granularity (spans per \texttt{<pause>}), and (iv) sensitivity to key hyperparameters.

\subsubsection{Controllability under external \texttt{<pause>} insertion.} \label{sec:ablation_pause_freq}
\begin{figure}[t]
  \centering
  \captionsetup{skip=3pt}
  \includegraphics[width=\linewidth]{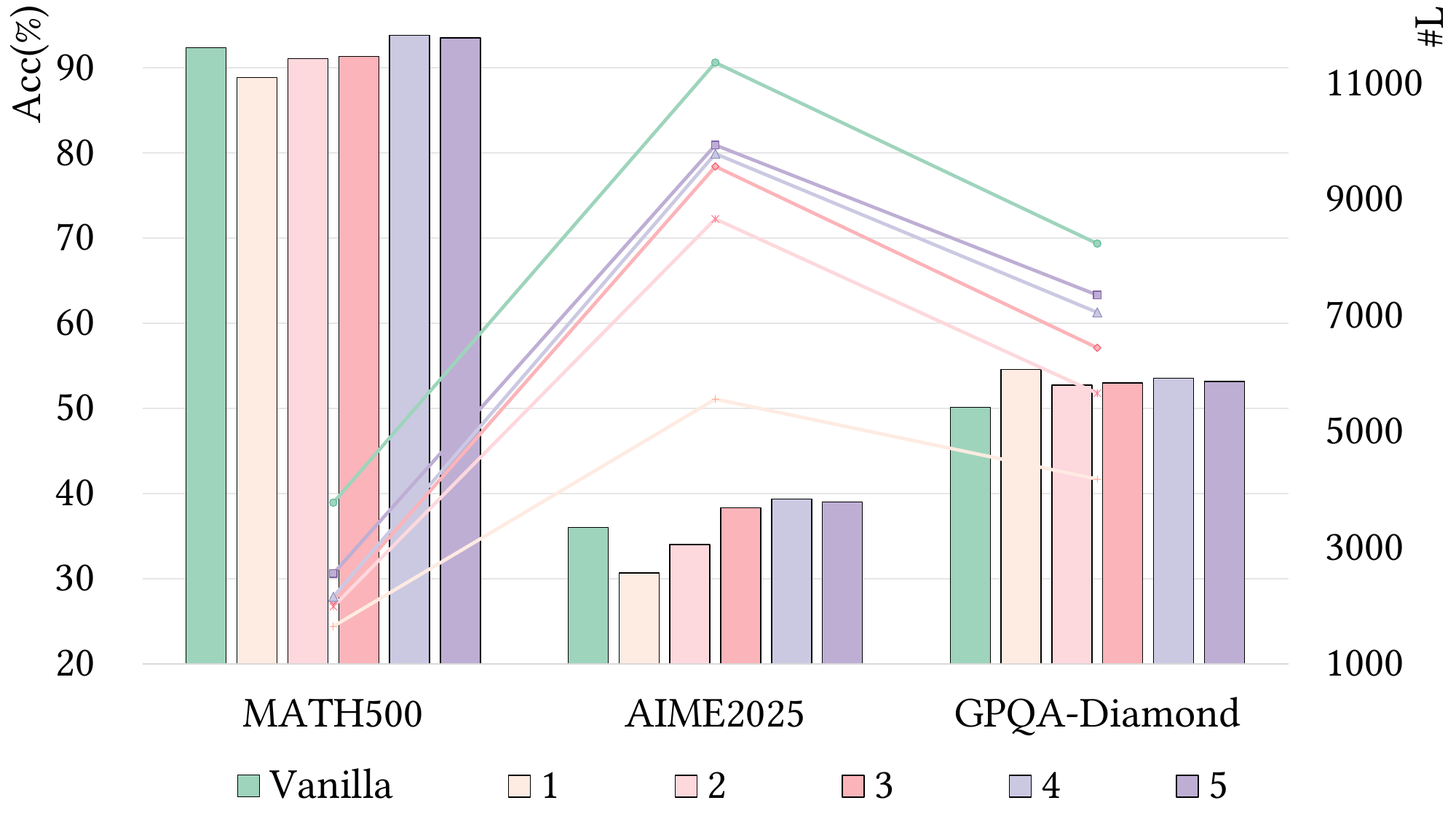}
  \caption{\textbf{Controllability under external \texttt{<pause>} insertion.}
  Bars report Pass@1 accuracy (\textbf{Acc}, left axis) and lines report output length (\textbf{\#L}, right axis) when inserting one \texttt{<pause>} after every $N$ spans within the \texttt{<think>} segment.}
  \label{fig:ablation_pause_freq}
\end{figure}

This experiment varies the \texttt{<pause>} insertion interval: within the \texttt{<think>} segment, we insert one \texttt{<pause>} after every $N$ explicit spans (span boundaries follow the blank-line rule in Section~\ref{sec:2_background}).
Figure~\ref{fig:ablation_pause_freq} shows that increasing insertion frequency (smaller $N$) generally reduces output length, confirming that external \texttt{<pause>} insertion provides a direct control over explicit verbosity.
Very dense insertion can reduce accuracy, reflecting the expected compression--performance tension.
Overall, external insertion provides predictable length control, with accuracy degrading only under very dense insertion.

\subsubsection{Alignment objective.}

\newcommand{\celllen}[2]{#1\raisebox{-0.35ex}{\tiny \textcolor{black!70}{#2}}}
\begin{table}[htbp]
\centering
\setlength{\tabcolsep}{1pt}
\renewcommand{\arraystretch}{1.08}

\resizebox{\linewidth}{!}{%
\begin{tabular}{lcccccc}
\toprule
\multirow{2}{*}{Method} &
\multicolumn{2}{c}{GSM8K} &
\multicolumn{2}{c}{AIME2025} &
\multicolumn{2}{c}{GPQA-Diamond} \\
\cmidrule(lr){2-3}\cmidrule(lr){4-5}\cmidrule(lr){6-7}
& Acc (\%) & \#L & Acc (\%) & \#L & Acc (\%) & \#L \\
\midrule
Vanilla
& 89.61 & 1316
& \underline{36.00} & 11357
& \underline{50.10} & 8238 \\

SPOT End\_KL
& 87.30 & \celllen{1204}{$_{-8.49\%}$}
& 21.33 & \celllen{10223}{$_{-9.99\%}$}
& 48.48 & \celllen{6910}{$_{-16.12\%}$} \\

SPOT MSE
& \underline{88.93} & \celllen{921}{$_{-29.99\%}$}
& 15.67 & \celllen{12785}{$_{+12.58\%}$}
& 43.43 & \celllen{8586}{$_{+4.22\%}$} \\

\rowcolor{mygray}
SPOT Sinkhorn
& \textbf{}{92.72} & \celllen{630}{$_{-52.13\%}$}
& \textbf{39.33} & \celllen{9568}{$_{-15.75\%}$}
& \textbf{54.55} & \celllen{4181}{$_{-49.25\%}$} \\
\bottomrule
\end{tabular}
}
\caption{Ablation on the alignment objective. Each \#L cell reports the absolute length with a subscript showing relative change vs.\ Vanilla.}
\label{tab:ablation_align_objective}
\end{table}

Table~\ref{tab:ablation_align_objective} shows that the alignment objective is critical.
We compare Sinkhorn OT with two heuristic alternatives that use the same frozen-head projection space.
\textbf{End\_KL} provides endpoint-only supervision by forcing the \texttt{<pause>} state to match the frozen-head decoding distribution of the \emph{last} teacher token in the span.
\textbf{MSE} uses a pooled target by averaging the teacher span into a single representation and regressing the \texttt{<pause>} state to this average, which ignores token-level structure within the span.

Replacing Sinkhorn OT with End\_KL or MSE leads to clear accuracy drops, with the largest degradation on the hardest math benchmark (AIME2025).
These alternatives can also weaken length control on difficult settings, yielding longer generations (e.g., MSE on AIME2025 and GPQA-Diamond).
In contrast, Sinkhorn OT achieves the most favorable accuracy--length profile across the three benchmarks, supporting its role as a robust many-to-one alignment between a single \texttt{<pause>} state and a variable-length teacher span.

\subsubsection{Spans per \texttt{<pause>}.}

\begin{table}[htbp]
\centering
\setlength{\tabcolsep}{2pt}
\renewcommand{\arraystretch}{1.10}

\resizebox{\linewidth}{!}{%
\begin{tabular}{lcc|cc|cc}
\toprule
\multirow{2}{*}{$G$} &
\multicolumn{2}{c|}{MATH500} &
\multicolumn{2}{c|}{AIME2025} &
\multicolumn{2}{c}{GPQA-Diamond} \\
\cmidrule(lr){2-3}\cmidrule(lr){4-5}\cmidrule(lr){6-7}
& Acc (\%) & \#L
& Acc (\%) & \#L
& Acc (\%) & \#L \\
\midrule
Vanilla
& 92.40 & 3775
& 36.00 & 11357
& 50.10 & 8238 \\
\midrule

1
& \textbf{93.80} & \celllen{2154}{$_{-42.95\%}$}
& \textbf{39.33} & \celllen{9568}{$_{-15.75\%}$}
& \textbf{54.55} & \celllen{4181}{$_{-49.25\%}$} \\

2
& \underline{90.62} & \celllen{1675}{$_{-55.64\%}$}
& \underline{32.67} & \celllen{9478}{$_{-16.54\%}$}
& \underline{48.49} & \celllen{3565}{$_{-56.73\%}$} \\

3
& 89.84 & \celllen{1606}{$_{-57.47\%}$}
& 30.67 & \celllen{9230}{$_{-18.72\%}$}
& 47.65 & \celllen{3426}{$_{-58.42\%}$} \\
\bottomrule
\end{tabular}%
}

\caption{\textbf{Spans per \texttt{<pause>}.} Each \texttt{<pause>} compresses $G$ consecutive reasoning spans.}
\label{tab:ablation_spp_3sets}
\end{table}

This ablation varies the compression granularity by letting each \texttt{<pause>} summarize $G$ consecutive reasoning spans, while keeping the maximum number of consecutive \texttt{<pause>} fixed to one.
As shown in Table~\ref{tab:ablation_spp_3sets}, increasing $G$ yields further token reduction on all three benchmarks, but introduces a clear accuracy degradation.
$G{=}1$ achieves the best overall trade-off, whereas $G{=}2$ or $3$ provides more aggressive compression at the cost of reliability, especially on harder benchmarks.

\subsubsection{Hyperparameter sensitivity analysis.}


\begin{table}[htbp]
\centering
\setlength{\tabcolsep}{2pt}
\renewcommand{\arraystretch}{1.08}
\begin{tabular}{lcccccc}
\toprule
\multirow{2}{*}{$\lambda$} &
\multicolumn{2}{c}{GSM8K} &
\multicolumn{2}{c}{AIME2025} &
\multicolumn{2}{c}{GPQA-Diamond} \\
\cmidrule(lr){2-3}\cmidrule(lr){4-5}\cmidrule(lr){6-7}
& Acc (\%) & \#L & Acc (\%) & \#L & Acc (\%) & \#L \\
\midrule
Vanilla
& 89.61 & 1316
& 36.00 & 11357
& 50.10 & 8238 \\
\midrule
0.0
& 86.66 & \celllen{1235}{-6.12\%}
& 24.00 & \celllen{11606}{+2.20\%}
& 34.34 & \celllen{8192}{-0.55\%} \\
0.2
& 88.57 & \celllen{1057}{-19.67\%}
& 28.00 & \celllen{10149}{-10.63\%}
& 47.47 & \celllen{6861}{-16.71\%} \\
0.4
& 89.23 & \celllen{1082}{-17.77\%}
& 29.67 & \celllen{9796}{-13.74\%}
& 47.98 & \celllen{6327}{-23.20\%} \\
0.6
& 90.00 & \celllen{969}{-26.35\%}
& 31.33 & \celllen{9621}{-15.28\%}
& 48.99 & \celllen{5870}{-28.74\%} \\
0.8
& \underline{91.67} & \celllen{822}{-37.57\%}
& \underline{37.33} & \celllen{10005}{-11.90\%}
& \underline{51.99} & \celllen{4061}{-50.70\%} \\
1.0
& \textbf{92.72} & \celllen{630}{-52.13\%}
& \textbf{39.33} & \celllen{9568}{-15.75\%}
& \textbf{54.55} & \celllen{4181}{-49.25\%} \\
\bottomrule
\end{tabular}
\caption{Sensitivity to the alignment weight $\lambda$.}
\label{tab:ablation_lambda}
\end{table}

As shown in Table~\ref{tab:ablation_lambda}, removing alignment ($\lambda=0$) leads to a clear accuracy drop, indicating that CE supervision alone is insufficient to reliably internalize span semantics.
Increasing $\lambda$ monotonically improves accuracy on all three benchmarks, while also reducing output length, with the best overall performance achieved at $\lambda=1.0$.
Moderate values (e.g., $\lambda=0.6$--$0.8$) already deliver substantial compression, but may underperform $\lambda=1.0$ on the most challenging settings.

\subsection{Interpretability Analysis}
Beyond task performance and generation length, this section evaluates the interpretability of \texttt{<pause>} from three complementary perspectives: (i) training-time alignment diagnostics on SpanDrop examples, (ii) LLM-as-a-Judge assessment of reasoning faithfulness under inference-time compression, and (iii) qualitative case studies under different external \texttt{<pause>} insertion patterns.

\subsubsection{Training-time Alignment Diagnostics}
This subsection examines what is encoded by \texttt{<pause>} states during Stage~I, using SpanDrop training examples where each \texttt{<pause>} is paired with a teacher span.
We track (i) the span-level Sinkhorn OT alignment loss in Eq.~\ref{eq:spot_align}, and (ii) a frozen-head top-$K$ coverage score measuring overlap between tokens decoded from the \texttt{<pause>} state and tokens appearing in the paired teacher span (Eq.~\ref{eq:topk_overlap}).

\noindent\textbf{Span-level OT alignment loss.}
We report the Sinkhorn-regularized OT value minimized by Eq.~\ref{eq:spot_align} (Algorithm~\ref{alg:spot_ot_align_compact}).
Lower values indicate that the projected \texttt{<pause>} representation is closer to the set of projected teacher token representations within the corresponding span.

\noindent\textbf{Frozen-head top-$K$ coverage.}
Let $h_{\texttt{<pause>}}$ denote the final-layer hidden state at a \texttt{<pause>} position.
We decode $h_{\texttt{<pause>}}$ with the frozen head (Eq.~\ref{eq:fh_prob}) and obtain the top-$K$ token set $\mathcal{T}_K(h_{\texttt{<pause>}})$ (Eq.~\ref{eq:topk_def}).
For the paired teacher span $S$, let $\mathcal{V}(S)$ be the set of unique token IDs in $S$ under the model tokenizer, after removing special tokens and trivial formatting tokens.
We define the normalized coverage as
\begin{equation}
\label{eq:topk_overlap}
\mathrm{coverage}(S)
\;=\;
\frac{\bigl|\mathcal{T}_{K}(h_{\texttt{<pause>}})\cap \mathcal{V}(S)\bigr|}
{\min\!\bigl(K,\;|\mathcal{V}(S)|\bigr)},
\qquad K=20.
\end{equation}

\begin{figure}[h]
  \centering
  \captionsetup{skip=3pt}
  \includegraphics[width=\linewidth]{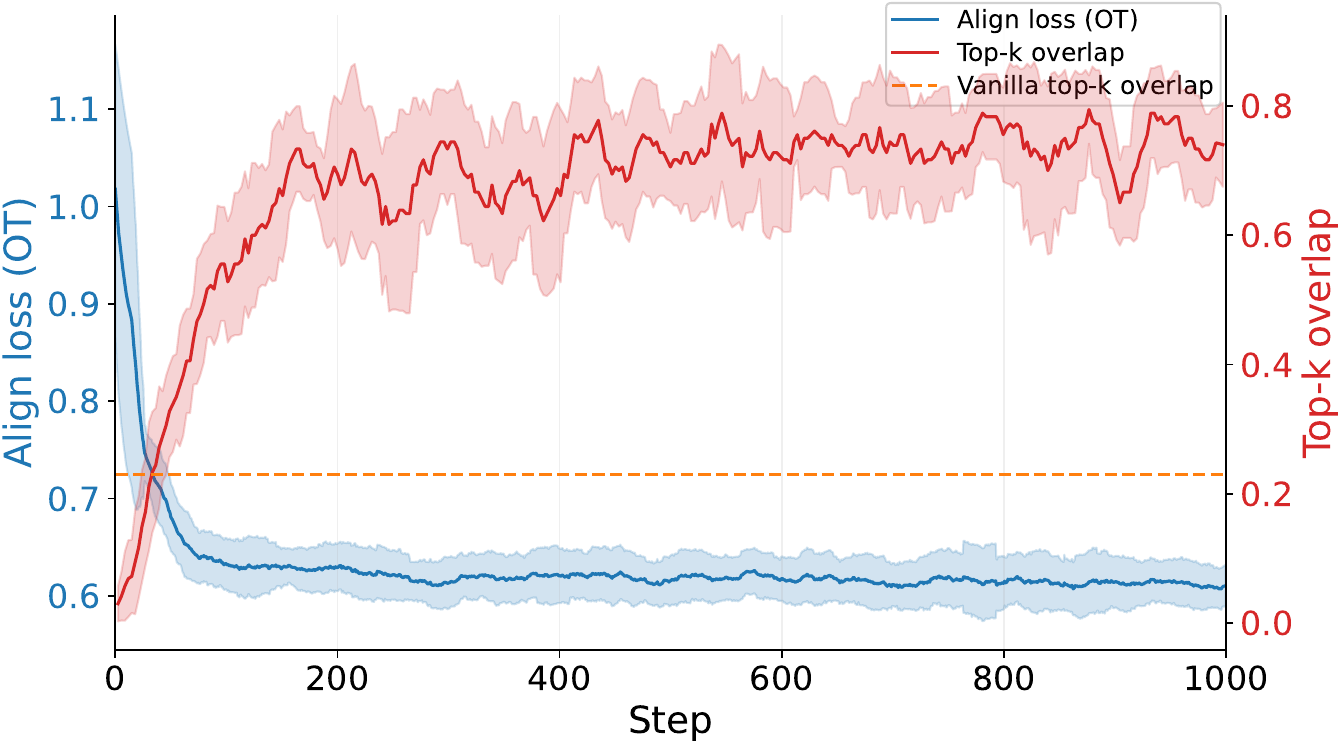}
  \caption{Training-time alignment diagnostics on SpanDrop examples.
  \textbf{Left axis:} span-level Sinkhorn OT alignment loss (lower is better).
  \textbf{Right axis:} frozen-head top-$K$ coverage between tokens decoded from the \texttt{<pause>} hidden state and the token set of the paired teacher span (Eq.~\ref{eq:topk_overlap}, $K{=}20$; higher is better).
  The dashed line denotes the vanilla backbone with externally inserted \texttt{<pause>} tokens evaluated under the same computation.}
  \label{fig:align_diagnostics}
\end{figure}

As a reference, we compute the same coverage score for the vanilla backbone with externally inserted \texttt{<pause>} tokens but without Stage~I training; the resulting baseline is shown in Figure~\ref{fig:align_diagnostics}.

Figure~\ref{fig:align_diagnostics} shows that the OT alignment loss decreases rapidly and then stabilizes, while the top-$K$ coverage rises substantially above the vanilla baseline.
Together, these trends suggest that Stage~I progressively induces \texttt{<pause>} states whose frozen-head decoding is increasingly consistent with the vocabulary content of their paired teacher spans.

\subsubsection{LLM-as-a-Judge Evaluation of Compressed Reasoning}

To further test whether externally inserted \texttt{<pause>} tokens act as a \emph{latent carrier} for non-trivial compression, we adopt an LLM-as-a-Judge protocol that scores text \emph{locally} around each \texttt{<pause>} boundary.
Under the same inference-time insertion schedule (Section~\ref{sec:inference}), we extract boundary pairs $(\texttt{PREVIOUS\_SPAN}, \texttt{NEXT\_SPAN})$, where spans are paragraphs in \texttt{<think>} (split by \texttt{\textbackslash n\textbackslash n}).

The judge assigns two 1--5 scores.
\textbf{pause\_utilization} measures whether \texttt{NEXT\_SPAN} relies on an intermediate conclusion \emph{not stated} in \texttt{PREVIOUS\_SPAN}, i.e., evidence of a meaningful jump across the \texttt{<pause>} token.
\textbf{local\_continuity} measures whether \texttt{NEXT\_SPAN} remains consistent with \texttt{PREVIOUS\_SPAN} (allowing a minimal bridge), penalizing contradictions or broken variable tracking.
We report mean scores and \textbf{Joint@4}, the fraction of boundaries where both scores are $\ge 4$ (Appendix~\ref{ap:llm_judge}).

\begin{table}[htbp]
\centering
\setlength{\tabcolsep}{6pt}
\renewcommand{\arraystretch}{1.10}
\resizebox{\linewidth}{!}{%
\begin{tabular}{lccc}
\toprule
Method
& \textbf{Local} $\uparrow$
& \textbf{Util.} $\uparrow$
& \textbf{Joint@4} $\uparrow$ (\%) \\
\midrule
Vanilla
& 3.59
& 1.71
& 17.2 \\
SPOT (End\_KL)
& 3.65
& \underline{3.20}
& \underline{29.7} \\
SPOT (MSE)
& \underline{3.98}
& 2.83
& 30.6 \\
\rowcolor{mygray}
SPOT (Sinkhorn)
& \textbf{4.85}
& \textbf{4.34}
& \textbf{83.6} \\
\bottomrule
\end{tabular}%
}
\caption{\textbf{LLM-as-a-Judge evaluation of \texttt{<pause>} boundaries.}
We report mean scores over all evaluated boundaries.
\textbf{Joint@4} is the percentage of boundaries where both \textbf{Local} and \textbf{Util.} are at least 4 (out of 5).}
\label{tab:llm_judge}
\end{table}

As shown in Table~\ref{tab:llm_judge}, SPOT improves \textbf{pause\_utilization} over the vanilla backbone while maintaining high \textbf{local\_continuity}, resulting in a much higher \textbf{Joint@4}.
This indicates that \texttt{<pause>} more often coincides with meaningful compression gaps under SPOT, without disrupting local coherence.

\subsubsection{Case Studies}
To illustrate the effect of \texttt{<pause>} on generation, we present representative examples that compare \texttt{<pause>}-augmented traces with the final solution.

\begin{figure}[h]
  \centering
  \captionsetup{skip=3pt}
  \includegraphics[width=\linewidth]{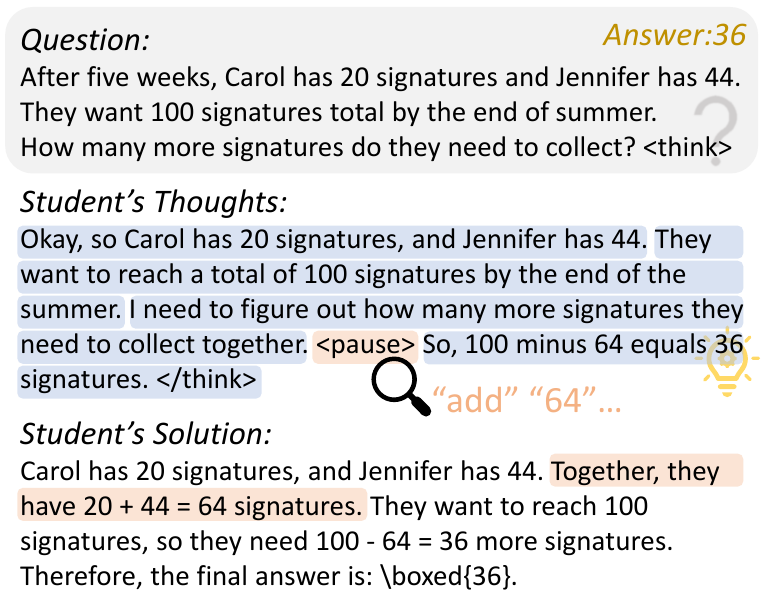}
  \caption{A successful \texttt{<pause>} compression example.}
  \label{fig:case}
\end{figure}

Figure~\ref{fig:case} shows a typical successful case: an intermediate aggregation is replaced by a single \texttt{<pause>}, yet the continuation remains consistent with the implied value ($20+44=64$) and yields the correct answer ($100-64=36$).
Additional examples (including GPQA-Diamond and a failure case) are provided in Appendix~\ref{ap:case_study}.

\section{Related Work} \label{sec:6_related}

\paragraph{Efficient explicit reasoning.}
A prominent direction improves reasoning while keeping intermediate steps textual, aiming to make the external traces \emph{more structured} and \emph{less wasteful}.
In modeling and prompting, prior work elicits higher-level decomposition or computation-oriented traces (e.g., abstraction or programs) and improves reliability by searching or aggregating multiple candidate rationales~\cite{zheng2024stepback,chen2023program,wang2023selfconsistency,yao2023tot}.
Complementary efforts focus on efficiency at inference time by controlling or shortening the generated trace---for instance, using concise guidance, difficulty-aware shortening, or explicit compute-budget control~\cite{tang2025concisehint,wu2025litecot,aytes2025sot,aggarwal2025l1,LLMTM}, and pruning low-utility parts via entropy signals or early-exit mechanisms~\cite{li2025stepentropy,pear2025phaseentropy,deer2025dynamicexit,wu2025cotpruning,DarkDistill,ToP,TokenPruning_kim}.
These approaches preserve human-readable reasoning, but their computation still scales with token-level decoding, making it difficult to substantially reduce cost without reducing available test-time compute.

\paragraph{Implicit and latent reasoning.}
Another line of work seeks to \emph{internalize} multi-step computation into hidden states, exposing only compact “thinking” interfaces.
One strand increases test-time compute with special or filler tokens before producing an answer, suggesting that additional decoding steps can provide computation even when the surface text is not informative~\cite{goyal2023pause,pfau2024dotbydot,herel2024thinkingtokens}.
Latent-CoT methods compress explicit rationales into continuous representations and operate in the latent space for subsequent reasoning~\cite{hao2024coconut,cheng2024ccot,shen2025codi}.
Recent systems further explore dynamic compression or latent-token generation to mitigate context growth while retaining reasoning ability~\cite{zhang2025lightthinker,colar2025openreview}.
To improve stability, hybrid paradigms interleave latent computation with explicit anchors~\cite{piao2025spiralthinker,li2025implicitreasoningsurvey}.
Despite this progress, existing designs often fall back to step-end matching or produce latent states with hard-to-audit semantics, leaving open the need for robust span-level alignment and interpretable latent thoughts.

\paragraph{Interpretability of latent thoughts.}
Interpretability is increasingly viewed as a prerequisite for deploying latent reasoning systems beyond controlled benchmarks.
While several methods attempt to decode latent thoughts into text, the resulting interpretations can be unstable when the latent vectors are not naturally compatible with the pretrained language head~\cite{shen2025codi,cheng2024ccot}.
A recent and influential direction represents latent thoughts as \emph{vocabulary-space superpositions}, making hidden states directly decodable as token distributions~\cite{deng2025latentsft}.

In this context, \textbf{SPOT} advances implicit reasoning by aligning compact latent thoughts to \emph{entire reasoning spans} and keeping them \emph{natively decodable under the frozen LM head}, while preserving a flexible interface that allows controllable use of implicit reasoning alongside optional explicit traces.

\section{Conclusion} \label{sec:8_conclusion}
In this work, we presented \textbf{SPOT}, a novel framework designed to compress explicit Chain-of-Thought (CoT) trajectories into compact latent \texttt{<pause>} tokens.
Crucially, SPOT preserves a flexible interface that facilitates the external insertion of tokens during inference.
By integrating span-level alignment with a frozen-head projection mechanism, our approach anchors each \texttt{<pause>} state to its corresponding teacher span within a vocabulary-induced embedding space, thereby ensuring that latent states remain natively decodable via the pretrained LM head.
Empirical evaluations across mathematical reasoning and out-of-domain science QA benchmarks demonstrate that SPOT significantly improves the trade-off between accuracy and generation length, while simultaneously supporting lightweight interpretability through frozen-head decoding.
Future research may explore generalizing the framework beyond heuristic blank-line segmentation by incorporating learnable or task-adaptive span boundaries; such advancements would enable more robust compression in complex domains, including planning and long-horizon decision-making.

\newpage

\bibliographystyle{ACM-Reference-Format}
\bibliography{reference}

\appendix
\section{Appendix} \label{sec:appendix}

\subsection{Alternative span granularities and insertion strategies} \label{ap:insertion}
SPOT is designed to decouple \texttt{<pause>} from any fixed interleaving template.
Unlike many implicit/hybrid reasoning frameworks that predefine a rigid alternation between latent tokens and explicit text (thereby enforcing a particular reasoning format), SPOT treats \texttt{<pause>} as an externally controllable interface: the model is trained to \emph{use} \texttt{<pause>} when it appears in the context, rather than being trained to emit it according to a fixed pattern.
This makes training non-intrusive with respect to the model’s native reasoning style, while enabling implicit computation to be activated at inference time by choosing where and how densely to insert \texttt{<pause>}.

\paragraph{Span boundary choices.}
In the main paper we define explicit spans using blank-line delimiters (\texttt{\textbackslash n\textbackslash n}), which match paragraph-level steps in DeepSeek-R1-style traces and yield a readable, tokenizer-independent segmentation.
More generally, the span boundary definition is modular: any deterministic rule that partitions the reasoning segment into consecutive text spans can be used.
For instance, one may adopt coarser paragraph-level spans, finer sentence-level spans, or heuristic spans based on formatting markers (e.g., numbered steps) when present.
Changing the granularity primarily affects what a single \texttt{<pause>} is asked to summarize and how frequently \texttt{<pause>} appears, but does not alter the core training objective or the inference interface.

\paragraph{Insertion strategies beyond spans.}
While our default inference-time schedule inserts \texttt{<pause>} after every $N$ spans for interpretability and stable control, SPOT also supports alternative insertion policies.
In addition to span-based insertion, \texttt{<pause>} can be placed at fixed token or character intervals, at dynamically chosen positions (e.g., after equation blocks), or with a prompt-dependent density schedule.
Empirically, we observe that these different insertion policies yield comparable accuracy--length trade-offs when matched to a similar overall \texttt{<pause>} density, suggesting that SPOT’s improvements are not tied to a single boundary definition or a single insertion heuristic.
This robustness aligns with the training goal of shaping \texttt{<pause>} into a general latent carrier of intermediate computation, rather than overfitting to a particular explicit/implicit alternation template.

\paragraph{Why we use span-based insertion in the main paper.}
We report span-based insertion as the default because it provides a clean, human-interpretable control knob (insert after every $N$ explicit steps), admits a tokenizer-independent implementation (operate on raw text), and minimally disrupts the readability of the remaining explicit reasoning.
Other policies (e.g., token-interval insertion) are fully compatible with SPOT, but can make the visible trace less structured and harder to inspect, even when task performance is similar.

\subsection{Benchmarks}
\label{ap:benchmarks}

\subsubsection{Benchmark Details} \label{ap:benchmark_details}
We evaluate on four math reasoning benchmarks and one out-of-domain (OOD) science QA benchmark.
Table~\ref{tab:benchmark_stats} summarizes dataset sizes and answer formats.

\noindent\textbf{GSM8K.}
GSM8K consists of grade-school math word problems with short numerical answers~\cite{cobbe2021gsm8k}.
We train on the GSM8K training split and report results on its official test split.

\noindent\textbf{MATH500.}
MATH is a challenging competition-level math dataset spanning multiple subjects (e.g., algebra, geometry, number theory)~\cite{hendrycks2021math}.
We report results on MATH500, a 500-problem evaluation subset commonly used to assess harder mathematical reasoning.
Compared with GSM8K, MATH500 typically requires longer, multi-step derivations.

\noindent\textbf{AIME 2024/2025.}
We evaluate on the American Invitational Mathematics Examination (AIME) problems from years 2024 and 2025.
AIME problems are competition-style and require producing a final integer answer in $[0,999]$ (as provided by the official problem sets and solutions).
These benchmarks further stress long-horizon reasoning and robustness under high difficulty.

\noindent\textbf{GPQA-Diamond.}
To test OOD generalization beyond math, we evaluate on the GPQA-Diamond subset~\cite{rein2023gpqa}, which contains the hardest graduate-level science questions in GPQA.

\begin{table}[h]
\centering
\small
\setlength{\tabcolsep}{4pt}
\renewcommand{\arraystretch}{1.08}
\begin{tabular}{lccc}
\toprule
Benchmark & Split & \#Instances & Answer format \\
\midrule
GSM8K~\cite{cobbe2021gsm8k} & train / test & 7473 / 1319 & \texttt{\textbackslash boxed\{\}} \\
MATH500~\cite{hendrycks2021math} & test & 500 & \texttt{\textbackslash boxed\{\}} \\
AIME 2024 & test & 30 & \texttt{\textbackslash boxed\{\}} \\
AIME 2025 & test & 30 & \texttt{\textbackslash boxed\{\}} \\
GPQA-Diamond~\cite{rein2023gpqa} & test & 198 & \texttt{Answer: \$LETTER} \\
\bottomrule
\end{tabular}
\caption{Benchmark statistics.}
\label{tab:benchmark_stats}
\end{table}

\subsection{Prompt Templates}
\label{sec:appendix_prompts}

\paragraph{Overview.}
For reproducibility, we present the prompts used in our evaluation.
Each prompt is composed of two parts:
(i) a \emph{method-specific instruction} that controls reasoning style, and
(ii) a \emph{task-specific format constraint} that standardizes the required final-answer format.
Unless otherwise specified, we apply the same task-specific format constraint across all methods for the same benchmark.
For baselines evaluated via their released implementations, we follow their default prompting and evaluation pipeline to avoid deviating from the intended setup.

\subsubsection{Method-specific instructions}
\label{ap:prompt_method}

\promptbox{Vanilla (standard CoT)}{Please reason step by step.}

\promptbox{SPOT (pause-compressed CoT)}{Solve the following math problem efficiently and clearly. You should provide concise and efficient Chain-of-Thought (CoT) reasoning. For any intermediate steps that are redundant, self-evident, or can be safely omitted without losing crucial information for the final solution, you must replace them with the special token <pause>, then continue seamlessly with the next logical step without restating previous content.}

\subsubsection{Task-specific format constraints}
\label{ap:prompt_task}

\promptbox{Math reasoning benchmarks (GSM8K/MATH500/AIME 2024/2025)}{The last line of your response should be of the following format: 'Therefore, the final answer is: \textbackslash boxed\{\}. I hope it is correct' (without quotes) where ANSWER is just the final number or expression that solves the problem.}

\promptbox{GPQA-Diamond (multiple-choice, Diamond subset)}{The last line of your response should be of the following format: 'Answer: \$LETTER' (without quotes) where LETTER is one of ABCD.}

\paragraph{Prompt assembly.}
Given an instance input (question, and choices if applicable), the final prompt is constructed by concatenating:
\emph{method-specific instruction} + \emph{task-specific format constraint} + \emph{instance content}.

\subsection{Baselines Details} \label{sec:appendix_baselines}
This section summarizes the core ideas of the baselines compared in our experiments.
Following the taxonomy in Section~\ref{sec:baselines}, we group methods into (i) approaches that reduce overthinking while keeping the reasoning trace explicitly produced, and (ii) approaches that shift a portion of intermediate computation into implicit/latent carriers.

\noindent \textbf{Efficient Explicit Reasoning Baselines} \label{ap:baseline_explicit}

\textbf{CCoT}~\cite{ccot}.
CCoT reduces the cost of explicit reasoning by compressing chain-of-thought into a more compact representation.
Compared with vanilla step-by-step traces, it targets shorter outputs by reducing redundancy in the produced reasoning while preserving the final answer fidelity.

\textbf{DEER}~\cite{deer2025dynamicexit}.
DEER improves efficiency via dynamic early exiting: it adaptively decides when to stop further reasoning/generation once sufficient confidence is reached.
The reasoning remains explicit when generated, but the method seeks to terminate decoding earlier to save tokens.

\textbf{ConciseHint}~\cite{tang2025concisehint}.
ConciseHint mitigates overthinking by injecting concise guidance during decoding, encouraging the model to produce shorter reasoning traces while preserving correctness.
It keeps the reasoning fully textual, but aims to reduce redundancy through controlled generation.

\textbf{Step Entropy}~\cite{li2025stepentropy}.
Step Entropy is a non-parametric compression strategy for explicit chain-of-thought.
It leverages uncertainty signals (e.g., token/step-level entropy) to identify and prune low-utility reasoning segments, thereby shortening the produced trace without changing the backbone parameters.

\textbf{L1}~\cite{aggarwal2025l1}.
L1 introduces an explicit regularization-based control of “thinking” to reduce unnecessary computation/verbosity.
By penalizing overly long or redundant reasoning behavior, it promotes more compact explicit outputs while aiming to maintain answer accuracy.

\noindent \textbf{Implicit Reasoning Baselines} \label{ap:baseline_latent}

\textbf{COCONUT}~\cite{hao2024coconut}.
COCONUT replaces parts of explicit chain-of-thought with continuous (non-text) latent representations, enabling the model to “think” in a hidden carrier and expose fewer reasoning tokens.
This shifts intermediate computation away from the visible output, aiming to maintain accuracy with significantly reduced explicit length.

\textbf{CODI}~\cite{shen2025codi}.
CODI distills explicit reasoning into internalized representations so that intermediate computation can be carried implicitly.
It encourages the student to reproduce the teacher's reasoning effect without explicitly emitting full chain-of-thought, reducing visible verbosity while preserving task performance.

\textbf{LightThinker}~\cite{zhang2025lightthinker}.
LightThinker targets “step-by-step compression” by reducing the number (or verbosity) of intermediate reasoning steps that appear in the output.
It encourages a compact reasoning process, typically by learning to compress or skip redundant steps while maintaining final-answer correctness.

\textbf{Latent-SFT}~\cite{deng2025latentsft}.
Latent-SFT trains models to perform reasoning in latent space via distillation-style supervision.
It internalizes intermediate computation into hidden representations (rather than explicit CoT tokens), and then generates a final response with fewer explicit reasoning tokens.

\subsection{Pseudo-code for Span-level OT Alignment} \label{ap:ot_pseudocode}
Algorithm~\ref{alg:spot_ot_align_compact} summarizes the span-level OT alignment loss used in Stage~I.
Algorithm~\ref{alg:sinkhorn_compact} gives the Sinkhorn solver with $\varepsilon$-scaling and numerical safeguards.

\begin{algorithm}[!ht]
\caption{Span-level OT alignment loss}
\label{alg:spot_ot_align_compact}
\KwIn{Dropped indices $\mathcal{D}$; student states $\{z_i\}$; teacher span states $\{H^{\text{tea}}(S_i)\}$; projection $\phi$; cap $T_{\max}$; \texttt{blur}, \texttt{scaling}; iters $L$; \texttt{normalize}; $\delta$.}
\KwOut{$\mathcal{L}_{\text{align}}$}
\If{$|\mathcal{D}|=0$}{\textbf{return} $0$\;}
$\mathcal{L}\leftarrow 0$\;
\ForEach{$i\in\mathcal{D}$}{
  $H \leftarrow H^{\text{tea}}(S_i)$\;
  \If{$|H|>T_{\max}$}{subsample $T_{\max}$ rows from $H$\;}
  $\tilde{z}\leftarrow \phi(z_i)$;\quad $\tilde{H}\leftarrow \phi(H)$\;
  \If{\texttt{normalize}}{
    $\tilde{z}\leftarrow \tilde{z}/(\|\tilde{z}\|_2+\delta)$;\;
    $\tilde{H}\leftarrow \tilde{H}\oslash(\|\tilde{H}\|_{2,\text{row}}+\delta)$\;
  }
  $C_{1,:}\leftarrow \|\tilde{z}-\tilde{H}\|_2^2$\;
  $\mathcal{L}\leftarrow \mathcal{L} + \textsc{SinkhornOTValue}(C,\texttt{blur},\texttt{scaling},L,\delta)$\;
}
$\mathcal{L}_{\text{align}}\leftarrow \mathcal{L}/|\mathcal{D}|$\;
\textbf{return} $\mathcal{L}_{\text{align}}$\;
\end{algorithm}

\begin{algorithm}[!ht]
\caption{Sinkhorn OT value with $\varepsilon$-scaling}
\label{alg:sinkhorn_compact}
\KwIn{$C\in\mathbb{R}_{+}^{m\times n}$; \texttt{blur}$>0$; \texttt{scaling}$\in(0,1]$; iters $L$; $\delta$.}
\KwOut{$\mathrm{OT}_{\varepsilon}(a,b;C)$ (approx.)}
$\varepsilon_{\min}\leftarrow \texttt{blur}^2$;\quad $\varepsilon \leftarrow \max(\varepsilon_{\min},\,\mathrm{median}(C))$\;
$a\leftarrow [1]\in\Delta^m$;\quad $b\leftarrow \mathbf{1}/n\in\Delta^n$\;
$u\leftarrow \mathbf{1}_m$;\quad $v\leftarrow \mathbf{1}_n$\;
\While{$\varepsilon > \varepsilon_{\min} + 10^{-12}$}{
  $K\leftarrow \exp(-C/\varepsilon)$\;
  \For{$\ell\leftarrow 1$ \KwTo $L$}{
    $u \leftarrow a \oslash (Kv+\delta)$;\;
    $v \leftarrow b \oslash (K^{\top}u+\delta)$\;
  }
  \If{\texttt{scaling}=1}{\textbf{break}\;}
  $\varepsilon \leftarrow \max(\varepsilon_{\min},\,\varepsilon\cdot\texttt{scaling})$\;
}
$\Pi \leftarrow \mathrm{diag}(u)\,K\,\mathrm{diag}(v)$;\quad $\Pi\leftarrow \max(\Pi,\delta)$\;
\textbf{return} $\langle \Pi, C\rangle - \varepsilon \sum_{i,j}\Pi_{ij}(\log\Pi_{ij}-1)$\;
\end{algorithm}

\FloatBarrier

\subsection{Additional case studies} \label{ap:case_study}
\paragraph{Case: GPQA-Diamond (full outputs).}
We include the full responses from the vanilla backbone and SPOT (with externally inserted \texttt{<pause>}) for qualitative comparison.

\casebox{Vanilla (full response)}{

Okay, so I've got this multiple-choice question about solving higher-dimensional heat equations using finite difference approximations and parallel splitting. Hmm, I'm a bit rusty on this, but let me try to think it through.

The question is about converting a sequential algorithm into a parallel one. The key factor here is what allows for parallel processing. I remember that when dealing with partial differential equations like the heat equation, finite difference methods are often used to discretize them. For higher dimensions, the equations can become quite complex.

The question mentions that the matrix exponential is approximated by a fractional approximation. Wait, fractional approximations... I think that refers to approximating the matrix exponential using methods like the Padé approximation or something similar. Oh right, matrix exponentials are used in solving systems of ODEs, which come from the discretization of PDEs. So when you have a PDE, you turn it into a system of ODEs, and the solution involves the matrix exponential.

Now, the key factor in converting a sequential algorithm to a parallel one. Parallel algorithms can take advantage of multiple processors or cores working on different parts of the problem simultaneously. So what makes this possible?

Option A: Complex roots of fractional approximation. Hmm. Fractional approximations could have complex roots, but I'm not sure how that relates to parallelism. Maybe the roots determine the stability or the approximation method, but not sure.

Option B: Existence of nonlocal boundary conditions. Nonlocal boundary conditions mean that the condition at a point depends on the solution elsewhere, not just adjacent points. That complicates things because you can't compute it locally. But does that enable parallelism? Maybe not directly. In fact, nonlocal conditions can make things harder to parallelize because you might need to wait for data from other parts of the domain.

Option C: Stability analysis. Stability is crucial for numerical methods. Without a stable method, you can't get convergence. But does that relate to parallelism? Stability analysis ensures that the method works, but it doesn't directly enable parallel processing. Unless the stability analysis determines how to split the problem into parallel parts, but I don't think that's the case here.

Option D: Linear partial fraction of fractional approximation. Oh, wait. The idea of partial fractions comes to mind. For example, if you have a rational function approximation (like a Padé approximant) of the matrix exponential, you might break it down into simpler fractions. When you do this, each term could correspond to a part of the problem that can be computed in parallel.

Like, suppose you have a matrix exponential approximated as a sum of terms, each of which can be computed independently. Then, each term could be handled by a different processor. So, the ability to represent the fractional approximation as a sum of linear terms (partial fractions) would allow for parallel computation of each term. That makes sense. So the key factor is the linear partial fraction decomposition.

So putting it all together, the correct answer should be D) Linear partial fraction of fractional approximation.
\texttt{</think>}
}

\casebox{SPOT (with externally inserted \texttt{<pause>})}{

Okay, so I've got this question about solving higher dimensional heat equations using finite difference methods, and converting a sequential algorithm into a parallel one. Hmm, let me think.

\texttt{<pause>}

The question mentions that the matrix exponential is approximated by a fractional approximation. Fractional approximations often involve expressing the exponential as a sum of simpler terms, maybe using partial fractions. Oh wait, that rings a bell. In parallel computing, if you can decompose the problem into parts that can be solved independently and then combined, it's more efficient. This is where methods like the Peaceman-Rachford or the Douglas-Rachford splitting come into play. These methods split the matrix into components that can be exponentiated separately.

\texttt{<pause>}

Looking at the options: option D says “Linear partial fraction of fractional approximation.” That sounds right because partial fractions would break the approximation into terms that can be handled separately.

\texttt{<pause>}

What about the other options? Option A is about complex roots. Not sure how that directly relates to parallelism. Option B talks about nonlocal boundary conditions, which I don't think is the key factor here. Option C is stability analysis, which is more about ensuring the method works rather than enabling parallelism.

\texttt{</think>}

The key factor in converting a sequential algorithm into a parallel one when using higher order finite difference approximations and parallel splitting lies in the ability to decompose the problem into independent tasks. This is achieved by approximating the matrix exponential using a fractional approximation that can be expressed as a linear partial fraction. This decomposition allows each term to be computed independently and in parallel, making the algorithm more efficient.

Answer: D
}

\noindent\textbf{Discussion.}
Compared with the vanilla completion, the \texttt{<pause>}-augmented trace retains the same overall response pattern (a brief problem restatement, option screening, and a final selection), but compresses the explicit narrative substantially.
In particular, several verbose explanatory sentences in the vanilla trace are replaced by a small number of \texttt{<pause>} markers, while the visible context around each marker remains locally coherent and still leads to the same final choice (D).
This example illustrates the intended use of \texttt{<pause>}: it reduces surface-level verbosity without forcing the model into a fixed template, and the final answer can remain supported by the surrounding explicit spans.

\paragraph{Case: GSM8K (failure under dense \texttt{<pause>} insertion).}
We next show a representative failure mode when \texttt{<pause>} is inserted too densely.
While the final answer can still be produced, the model may exhibit restart/repetition behavior and irregular \texttt{<pause>}/\texttt{<think>} boundaries, which increases verbosity and weakens local coherence.

\casebox{SPOT (dense \texttt{<pause>} insertion; full response)}{

Okay, so I need to figure out how many days the teacher can use a piece of chalk before he has to recycle it. Let me try to break this down step by step.

<pause><think>

First, I need to determine how much chalk is left after the teacher uses 45\% of a 5-inch piece on Monday. 

If the teacher uses 45\% of 5 inches, that's 0.45 * 5 = 2.25 inches used. 

Subtracting that from the original length, 5 - 2.25 = 2.75 inches remaining.

<pause><think>

Okay, so the teacher started with a 5-inch chalk piece on Monday. He used 45\% of it, which is 2.25 inches. That leaves 2.75 inches.

<pause>

Now, the teacher wants to use only 20\% of the chalk each day after Monday. 

20\% of 5 inches is 0.2 * 5 = 1 inch per day.

<pause>

Each day, the teacher uses 1 inch of chalk. Starting with 2.75 inches, I need to determine how many full days he can write before the chalk is less than 2 inches, at which point he'll have to recycle it.

<pause>

Since the teacher recycles when the chalk is smaller than 2 inches, he can use the chalk until it's just below 2 inches. 

Starting with 2.75 inches:

<pause>

On Day 1: 2.75 - 1 = 1.75 inches remaining. 

Since 1.75 inches is less than 2 inches, the teacher will have to recycle the chalk after Day 1.

<pause>

So, the teacher can use the chalk for 1 full day before needing to recycle it.

<pause>

The teacher can use the chalk for 1 day before it needs to be recycled.

</think>

}

\noindent\textbf{Discussion.}
This example highlights a typical failure mode under very dense external \texttt{<pause>} insertion.
Instead of suppressing redundant text, the model repeats intermediate computations and restates the same reasoning in multiple formats (free-form sentences and a numbered derivation), leading to a longer and less coherent trace.
Such restart/repetition behavior can reduce local continuity and obscure the intended compression effect, even when the final prediction remains plausible.
This motivates the controllability analysis in Section~\ref{sec:ablation_pause_freq} and the Stage~II stabilization, which reduces sensitivity to \texttt{<pause>} placement and density.

\subsection{LLM Judge Details} \label{ap:llm_judge}
To assess the faithfulness of compressed reasoning produced under inference-time external \texttt{<pause>} insertion, we use an LLM-as-a-Judge protocol.
The judge is instructed to focus on (i) local transition coherence across the \texttt{<pause>} boundary and (ii) whether \texttt{NEXT\_SPAN} relies on an intermediate conclusion \emph{not stated} in the \texttt{PREVIOUS\_SPAN} (i.e., evidence of meaningful compression), while strictly avoiding solving the problem or hallucinating missing steps.
Box~\ref{box:judge_prompt} shows the full prompt used in our evaluation.

\begin{tcolorbox}[judgebox, title={LLM-as-a-Judge prompt used in our evaluation.}, breakable, before upper=\raggedright\sloppy]
\label{box:judge_prompt}
\footnotesize\ttfamily

<System Prompt>\par
You are an evaluator of compressed reasoning quality around externally inserted \texttt{<pause>} markers.\par
Your role is to assess local continuity and pause utilization only.\par
Do not solve the problem. Do not rewrite the reasoning. Do not assume hidden steps beyond the provided text.\par
Return only a valid JSON object.\par

\medskip
<User Prompt>\par
You will evaluate one compressed reasoning instance.\par

\medskip
Scoring rubric (1--5):\par
\medskip

1) local\_continuity\par
5: \texttt{NEXT\_SPAN} follows smoothly from \texttt{PREVIOUS\_SPAN}; variable tracking is consistent; no contradictions.\par
4: Minor gap, but the intended bridge is clear and recoverable from the provided text.\par
3: Noticeable gap; at least one important bridge is ambiguous.\par
2: Major gap; \texttt{NEXT\_SPAN} does not follow without substantial missing reasoning.\par
1: Contradictory, incoherent, or non-sequitur transition.\par

\medskip
2) pause\_utilization\par
5: Clear evidence that \texttt{NEXT\_SPAN} uses an intermediate conclusion not explicitly stated in \texttt{PREVIOUS\_SPAN} (a meaningful ``jump'' across \texttt{<pause>}).\par
4: Some evidence of an unstated intermediate conclusion; the jump is plausible from context.\par
3: Weak/ambiguous evidence; \texttt{NEXT\_SPAN} could be derived without relying on an unstated intermediate result.\par
2: Little evidence of a jump; \texttt{NEXT\_SPAN} is mostly a straightforward continuation or minor rewrite.\par
1: No evidence of utilization; \texttt{<pause>} appears unnecessary (e.g., repetition, filler, or restating the same point).\par

\medskip
Constraints:\par
- Judge strictly from provided text.\par
- Do not credit unobserved ``implicit'' steps unless trivially inferable.\par
- Penalize restart/repetition/filler behavior when it harms coherence.\par
- Keep rationale concise and specific.\par

\medskip
Return ONLY JSON with exactly this schema:\par
\{ \par
\ \ "local\_continuity": 1,\par
\ \ "pause\_utilization": 1,\par
\ \ "rationale": "",\par
\ \ "missing\_step": "",\par
\ \ "error\_type": "none",\par
\ \ "confidence": 0.0\par
\}\par

\medskip
Where:\par
- rationale: short explanation for assigned scores ($\le$ 80 words).\par
- missing\_step: brief missing bridge if any, else "".\par
- error\_type: one of\par
\ \ ["none", "minor\_gap", "major\_gap", "contradiction",
   "restart", "repetition", "irrelevant", "format\_issue",
   "insufficient\_context"].\par
- confidence: real number in [0,1].\par

\medskip
[PROBLEM]\par
\{question\}\par

\medskip
[COMPRESSED\_TRACE]\par
\{compressed\_trace\}\par

\medskip
[PREVIOUS\_SPAN]\par
\{step\_before\}\par

\medskip
[NEXT\_SPAN]\par
\{step\_after\}\par

\medskip
[FINAL\_ANSWER]\par
\{final\_answer\}\par
\end{tcolorbox}

\paragraph{Case: LLM-as-a-Judge boundary example.}
We provide an example boundary pair extracted from a \texttt{<think>} segment, together with the judge outputs.
The \texttt{NEXT\_SPAN} jumps to a new derived quantity that relies on an intermediate result that is not explicitly stated in \texttt{PREVIOUS\_SPAN}, illustrating a high \textbf{pause\_utilization} score while remaining locally coherent.

\casebox{PREVIOUS\_SPAN}{

Since the first train departs at 30 minutes, the second arrives at $30 + 30 = 60$ minutes from now.

The second train stays for a quarter of the time the first train stayed. The first train stayed for 20 minutes, so a quarter of that is 5 minutes. So, the second train departs 5 minutes after it arrives, which would be at $60 + 5 = 65$ minutes from now.

}

\casebox{NEXT\_SPAN}{

The fourth train arrives 20 minutes after the third train leaves. So, $125 + 20 = 145$ minutes from now. Therefore, Paul has to wait until 145 minutes for the fourth train.

}

\casebox{Judge output (GPT-4o-mini)}{

\texttt{"pause\_utilization": 5,} \\
\texttt{"local\_continuity": 5,} \\
\texttt{"rationale": "NEXT\_SPAN directly uses an unstated intermediate time (the third train leaves at 125 minutes) to compute the fourth train arrival at 145, indicating a clear jump across the boundary. The continuation is consistent once this minimal bridge is supplied.",} \\
\texttt{"missing\_step": "Derive the third-train time: the third train arrives an hour after the second leaves, so $65 + 60 = 125$ minutes, and it departs immediately.",} \\
\texttt{"error\_type": "none",} \\
\texttt{"confidence": 0.9 }

}

\noindent\textbf{Discussion.}
This boundary exhibits a non-trivial jump: \texttt{NEXT\_SPAN} introduces the computation $125{+}20$ without explicitly stating how $125$ is obtained from \texttt{PREVIOUS\_SPAN}.
At the same time, the transition remains logically consistent with a single short bridge (computing the third-train time), aligning with our goal of using \texttt{<pause>} as a latent carrier that enables compression without introducing contradictions.

\paragraph{Case: LLM-as-a-Judge negative example (redundant \texttt{<pause>}).}
We also include a negative case where the judge LLM assigns \textbf{low pause\_utilization}.
Here, \texttt{NEXT\_SPAN} largely restates and instantiates what is already explicit in \texttt{PREVIOUS\_SPAN}, indicating that the inserted \texttt{<pause>} does not correspond to a meaningful compression jump.

\casebox{PREVIOUS\_SPAN}{

But in math problems, usually, ``times older'' means multiplied by that number. So, if someone is two times older than another, they're twice as old. So, Ben's age would be 2 times Suzy's age.

}

\casebox{NEXT\_SPAN}{

So Ben's age, let's call it $B$, is $B = 2 \cdot S$. Since $S$ is 1, that would make $B = 2 \cdot 1 = 2$. So Ben's iPhone is 2 years old.

}

\casebox{Judge output (GPT-4o-mini)}{

\texttt{"pause\_utilization": 1,} \\
\texttt{"local\_continuity": 5,} \\
\texttt{"rationale": "The NEXT\_SPAN does not introduce any new items or conclusions that were not already present in the PREVIOUS\_SPAN. It continues the discussion about Ben's age without adding new information, making the <pause> redundant.",} \\
\texttt{"missing\_step": "",} \\
\texttt{"error\_type": "none",} \\
\texttt{"confidence": 0.90}

}

\noindent\textbf{Discussion.}
In this boundary, the reasoning transition is fully explicit: \texttt{PREVIOUS\_SPAN} already states the key relationship (Ben is twice Suzy), and \texttt{NEXT\_SPAN} simply writes it as an equation and plugs in the given value.
Accordingly, the judge assigns high \textbf{local\_continuity} but minimal \textbf{pause\_utilization}, reflecting that the \texttt{<pause>} marker does not align with a non-trivial compression gap in this instance.

\end{document}